\documentclass[12pt]{article}

\usepackage{arxiv}

\usepackage{xcolor, etoolbox}

\usepackage{hyperref}

\hypersetup{
   colorlinks=true,
   urlbordercolor={1 1 1},
   citecolor={green}
}

\usepackage[utf8]{inputenc} 
\usepackage[T1]{fontenc}    
\usepackage{hyperref}       
\usepackage[left]{lineno}

\usepackage{url}            
\makeatletter
\g@addto@macro{\UrlBreaks}{\UrlOrds}
\makeatother

\usepackage{booktabs}       
\usepackage{amsfonts}       
\usepackage{nicefrac}       
\usepackage{microtype}      
\usepackage{lipsum}
\usepackage{amsmath}
\usepackage{amsfonts}
\usepackage{relsize}

\usepackage{amsmath}
\usepackage{amstext}
\usepackage{amssymb}

\usepackage{comment}
\usepackage{listings}

\usepackage{graphicx}
\usepackage{subcaption}
\usepackage{epstopdf}

\usepackage{multirow}
\usepackage{graphicx}
\usepackage{tabulary,capt-of}
\usepackage{slashbox}
\usepackage{adjustbox}
\usepackage{diagbox}
\usepackage{algorithm}

\usepackage{amsmath}
\usepackage{amsfonts}
\usepackage{algpseudocode}
\usepackage{multicol}
\usepackage{caption}

\usepackage{mathtools}
\newtagform{blue}{\color{red}(}{)}
\newtagform{redandblue}[\textcolor{red}]{\color{red}(}{)}

\usetagform{blue}

\newtheorem{theorem}{Theorem}

\newtheorem{rem}[theorem]{Remark}
\newenvironment{remark}{\rem\rm}{\endrem}

\newcounter{unnumber}

\usepackage{multirow}
\usepackage{colortbl}
\usepackage{multicol}

\title{New optimization algorithms for neural network training using operator splitting techniques}

\author{
  Cristian Daniel Alecsa\thanks{ \textbf{ Corresponding author. This work was supported by a grant of Ministry of Research and Innovation,
CNCS - UEFISCDI, project number PN-III- P1-1.1-TE-2016 - 0266, within PNCDI
III.}} \\
  Tiberiu Popoviciu Institute of Numerical Analysis\\
  Romanian Academy\\
  Cluj-Napoca, Romania, RO-400320 \\
  \texttt{cristian.alecsa@ictp.acad.ro} \\ \\
  Romanian Institute of Science and Technology \\
  Cluj-Napoca, Romania, RO-400022 \\
  \texttt{alecsa@rist.ro} \\ \\
   \And
 Titus Pin\c ta \\
  Mathematical Institute\\
  University of Oxford\\
  Oxford, England, OX2 6GG \\
  \texttt{Titus.Pinta@maths.ox.ac.uk} \\
  	\And
  Imre Boros \\
  Department of Mathematics\\
  Babe\c s-Bolyai University\\
  Cluj-Napoca, Romania, RO-400084 \\
  \texttt{imre.boros@math.ubbcluj.ro} \\ \\
  Tiberiu Popoviciu Institute of Numerical Analysis\\
  Romanian Academy\\
  Cluj-Napoca, Romania, RO-400320 \\
  \texttt{boros.imre@ictp.acad.ro}	
}
\begin{document}
\maketitle

\begin{abstract}
In the following paper we present a new type of optimization algorithms adapted for neural network training. These algorithms are based upon sequential operator splitting technique for some associated dynamical systems. Furthermore, we investigate through numerical simulations the empirical rate of convergence of these iterative schemes toward a local minimum of the loss function, with some suitable choices of the underlying hyper-parameters. We validate the convergence of these optimizers using the results of the accuracy and of the loss function on the MNIST, MNIST-Fashion and CIFAR 10 classification datasets.
\end{abstract}

\section{Introduction}\label{Introduction_Section}

The purpose of the present section is to revisit the basic algorithms that are used for unconstrained optimization problems. We focus our attention on the empirical order of convergence in the setting of deep and convolution neural networks. The rigorous mathematical proofs for the convex and non-convex cases will be given in a follow-up article. 
From now on, the objective function will be denoted by $f$, the underlying Lipschitz constant by $L$ and the step-size of the discretization scheme by $h$. Firstly, we recall that the best known numerical algorithm used in the setting of unconstrained optimization problems is the gradient descent scheme (see \cite{Nocedal}):

\begin{align}\label{GD}
u_{n+1} = u_{n} - h \nabla f(u_n).
\end{align}

This simple algorithm is in fact the explicit forward Euler method and it represents a proper numerical discretization for the following continuous dynamical system: 

\begin{align}\label{GD-DiffSyst}
\dot{u}(t) = -\nabla f(u(t)).
\end{align}

On the other hand, Polyak (see \cite{Polyak}) introduced a new type of accelerated gradient-type method based upon the idea of an inertial term. This is a two-step method that is heavily used in optimization applications. Polyak's momentum algorithm with constant coefficients is the following:

\begin{align}\label{Plk-Const}
\begin{cases}
u_{n+1} = y_n - h \nabla f(u_n) \\
y_n = u_n + \gamma (u_n - u_{n-1}).
\end{cases}
\end{align}

From \cite{Ruder} we remind that the gradient descent and the stochastic variant of gradient descent, namely SGD (see \cite{Bottou}  on Page 2) have trouble navigating landscapes that differ from one dimension to another. The momentum method algorithm \ref{Plk-Const} is used in order to damp the oscillations of the one-step methods like gradient descent and also in order to prevent the divergence of these type of algorithms. In the optimization literature, the case of the Polyak momentum algorithm with non-constant coefficients is thoroughly studied in the framework of both convex and non-convex minimization problems. This is given as follows:

\begin{align}\label{Plk-NonConst}
\begin{cases}
u_{n+1} = y_n - \beta_n \nabla f(u_n) \\
y_n = u_n + \alpha_n (u_n - u_{n-1}),
\end{cases}
\end{align}

where $\alpha_n \in [0,1)$ and $\beta_n > 0$. Quite interestingly, in \cite{Sun}, the authors proved an $\mathcal{O} \left( \tfrac{1}{n} \right)$ theoretical convergence rate, under the assumption that the objective function is coercive, $(\alpha_n)_{n \in \mathbb{N}}$ is a decreasing sequence and the step-size parameter satisfies $\beta_n = \tfrac{2(1-\alpha_n) c}{L}$, for a fixed element $c$ from $(0,1)$.

Also, the most important optimization algorithm is the inertial algorithm that was developed by Y. Nesterov in \cite{Nesterov}. This accelerated inertial scheme has the following form:

\begin{align}\label{NaG}
\begin{cases}
u_{n+1} = y_n - s \nabla f(y_n) \\
y_n = u_n + \beta_n (u_n - u_{n-1}).
\end{cases}
\end{align}

Moreover, Su, Boyd and Cand\' es \cite{Boyd} showed that the underlying dynamical system of \ref{NaG} is the following second-order differential system, with non-constant gradient damping:
 
\begin{align}\label{NaG-DiffSyst}
\ddot{u}(t) + \dfrac{3}{t} \dot{u}(t) = -\nabla f(u(t)).
\end{align}

It is known (see \cite{Ruder}) that \ref{NaG} is used in order to update the iteration values $u_n$ using the inertial coefficients. This will give us a practical way in order to approximate the next values of the parameters, using future updated values. Also, in machine learning applications, the inertial term $\beta_n$ is taken to be a constant with an approximate value between $0.5$ and $0.9$. \\
Further, it is known that \ref{NaG} exhibits an $\mathcal{O} \left( \tfrac{1}{n^2} \right)$ convergence rate, under the strong assumption of convexity of the objective function $f$, when $s \leq \tfrac{1}{L}$. Last but not least, we recall that another competitive forward-backward type algorithm was proposed in \cite{Bot} by Bo\c t, Csetnek and L\' aszl\' o. It is worth noticing that this algorithm is an extension of the Polyak's method if the nonsmooth term vanishes, in the setting of the non-convexity of the objective function. Finally, to wrap this introductory section up, we remind that in order to obtain the convergence rates of optimization schemes in the framework of non-convex minimization problems, the regularization function (which is, in fact, a discrete Lyapunov function) must obey the KL property (e.g. see \cite{Bot}).

\subsection{Some notes on Nesterov's accelerated method}\label{Nesterov_SubSection}

In this subsection we consider some explanations regarding the asymptotic behavior of Nesterov's accelerated gradient method. We emphasize the importance of the inertial damping sequence $(\beta_n)_{n \in \mathbb{N}}$ that converges to $1$ as $n$ goes to $\infty$ and as the underlying step-size is fixed. Moreover, the additional iteration sequence $(y_n)_{n \in \mathbb{N}}$ that, for every $n \in \mathbb{N}$, is similar to a predictor step in the direction of the critical points, can be written with respect to the discretized velocity. This key idea will be further used in the algorithms that we shall consider.

\begin{remark}\label{R1}{\ \\}
In \ref{NaG}, the momentum sequence $(\beta_n)_{n \in \mathbb{N}}$ has the general term defined as $\beta_n = \tfrac{n}{n+3}$ or  $\beta_n = \tfrac{n-1}{n+2}$, for $n \geq 1$  (for simplicity, we consider the last one). The constant term $s$ is taken to be $h^2$, where $h$ is the step-size of the numerical discretization. Furthermore, it is known that the \ref{NaG} algorithm can be considered as a numerical discretization based upon central and forward finite differences of the continuous counterpart \ref{NaG-DiffSyst}. In this case, one can write \ref{NaG} in the following form:

\begin{align*}
\dfrac{u_{n+1} - 2u_{n} + u_{n-1}}{h^2} + \dfrac{1-\beta_n}{h} \cdot \dfrac{u_n - u_{n-1}}{h} = - \nabla f(y_n) \ .
\end{align*}

From a numerical point of view, taking $t = nh$, i.e. $t = n \sqrt{s}$, one obtains that 

$$ \dfrac{1-\beta_n}{h} = \dfrac{3}{t+2h} \ . $$

Taking the limit as the step-size $h$ goes to $0$ for the fixed value of $t$, we obtain that 

$$\lim\limits_{h \to 0} \dfrac{1-\beta_n}{h} = \dfrac{3}{t} \ .$$

We are to highlight that the above limit will play a key role in the development of the splitting optimization algorithms that we will introduce after this section. Moreover, the inertial parameter $\beta_n$ converges to $1$ as $n$ goes to $+\infty$, since, with the above consideration, $\beta_n = \tfrac{t-h}{t+2h}$, as $n \geq 1$. \\
We can rewrite the \ref{NaG} algorithm in a more intuitive form, that is based on the concept of velocity of the underlying second order differential system. One can see that the dynamical system \ref{NaG-DiffSyst} can be written as

\begin{align}\label{NaG-ContSyst}
\begin{cases}
\dot{u}(t) = v(t) \\ 
\dot{v}(t) = - \dfrac{3}{t} v(t) - \nabla f(u(t)).
\end{cases}
\end{align}

Relating to \ref{NaG-ContSyst}, the Euler-type discretization leads to the following numerical algorithm:

\begin{align*}
\begin{cases}
\dfrac{u_{n+1} - u_{n}}{h} = v_{n+1} \\
\dfrac{v_{n+1}-v_{n}}{h} = - \dfrac{v_n (1-\beta_n)}{h} - \nabla f(y_n),
\end{cases}
\end{align*}

where the sequence $(y_n)_{n \in \mathbb{N}}$ will depend on the velocity term $v_n$, for $n \in \mathbb{N}$.

After some algebratic manipulations, the \ref{NaG} algorithm can be written in the following manner:

\begin{align}\label{Nag-DiscreteSyst}
\begin{cases}
u_{n+1} = u_n + h v_{n+1} \\
v_{n+1} = \beta_n v_n - h \nabla f(y_n).
\end{cases}
\end{align}

In the case of \ref{NaG}, the auxiliary iteration $y_n = u_n + \beta_n (u_n - u_{n-1})$ will lead to $y_n = u_n + h \beta_n v_n$, and consequently the optimization algorithms \ref{NaG} and \ref{Nag-DiscreteSyst} are mathematically equivalent. This means that the momentum sequence $(y_n)_{n \in \mathbb{N}}$ can be defined as a linear combination of the primary iteration $u_n$ and the underpinning velocity $v_n$. 

\end{remark}

\subsection{  The organization of the paper and the key ingredients of the splitting techniques }\label{Organization_SubSection}

In this subsection we briefly present the main ideas concerning the concept of operator splitting. Furthermore, we will explain in detail the role of each section and subsection, respectively. At the same time, we shall infer the role of the splitting methods in optimization problems, where we emphasize the most important mathematical details that are used in the derivation of the formulas concerning the sequential splitting discretizations. The paper is organized as follows: in the present section, namely Section \ref{Introduction_Section}, we have presented a brief introduction into the most used optimization algorithms along with their dynamical systems. Furthermore, in Subsection \ref{Nesterov_SubSection}, we have considered a crucial remark concerned with the asymptotical behavior of Nesterov method \ref{NaG}. This observation is related to the fact that the additional iteration $y_n$ can be written as $y_n = u_n + h \beta_n v_n$, where $v_n$ is the usual velocity at the discrete time $t = nh$. At the same discretization time $t$, the inertial sequence $(\beta_n)_{n \in \mathbb{N}}$ is given by $\beta_n = (t - h)/(t + 2h)$, so it converges assymptotically to $1$ as time grows to infinity. This observations will be the main ingredients in constructing sequential splitting-based optimization algorithms. On the other hand, in Section \ref{Methods_Section}, we consider the methods and techniques that represent the backbone of our research. These are based on the idea that an abstract evolution equation of the form $\dot{X}(t) = (A+B) X(t)$ can be decomposed into two subproblems, i.e. $\dot{X}(t) = A X(t)$ and $\dot{X}(t) = B X(t)$, respectively. This leads to the natural idea that, instead of using a numerical method for the full dynamical system, we employ two discretizations, one for each subproblem. So, if $X(t) = (u(t), v(t))$, then at the discrete time $t = nh$, the first discretization operator maps $(u_n, v_n)$ into $(u_{n+1/2}, v_{n+1/2})$, and the second discretization operator that is in fact related to the second subproblem maps $(u_{n+1/2}, v_{n+1/2})$ into $(u_{n+1}, v_{n+1})$. It is known that this simple approach of splitting a vector field into two subvector fields is keen on the idea of separating the parts of the dynamical systems that have different physical interpretations. Now, in Subsection \ref{SSA1_SubSection} and in Subsection \ref{SSA2_SubSection} we introduce two completely new optimization algorithms. The key idea is that each of them is a discretization, as time is large enough, to an assymptotical evolution equation in the finite dimensional space $\mathbb{R}^N$, namely \ref{SeqSplittI-DiffSyst}. We shall explain briefly the structure of each of these numerical methods. Regarding Subsection \ref{SSA1_SubSection}, for the first splitting algorithm, we consider the following dynamical system:
\begin{align}\label{SSA1-DynSys}
\ddot{u}(t) + \left[ \delta(t) - 2 \dfrac{\dot{\delta}(t)}{\delta(t)} \right]  \dot{u} (t) = - \delta^2 (t) \nabla f(u(t)).
\end{align}
Then, using the notation $\dot{u}(t) = \delta^2(t) v(t)$, we consider the following attached subproblems:
\begin{align*}
\begin{cases}
\dot{u}(t) = 0 \\
\dot{v}(t) = - \delta(t) v(t)
\end{cases}
\text{ and } \quad
\begin{cases}
\dot{u}(t) = \delta^2 (t) v(t) \\
\dot{v}(t) = -  \nabla f(u(t)).
\end{cases}
\end{align*}
Then, using on the first subproblem \ref{Pb-1-discrete} and on the second one \ref{Pb-2-discrete-original-I}, we obtain our first splitting algorithm, but, as we shall see, with some modifications on the velocity. The same technique is applied to a different dynamical system, namely:
\begin{align}\label{SSA2-DynSys}
\ddot{u}(t) + \delta (t) \dot{u} (t) = - \nabla f(u(t)).
\end{align}
Now, for the full dynamical system we use the notation $\dot{u}(t) = v(t)$. As before, we have two dynamical systems that are represented by the sum of two vector fields, such that this sum is equal to the original vector field of \ref{SSA2-DynSys}. This is equivalent to
\begin{align*}
\begin{cases}
\dot{u}(t) = 0 \\
\dot{v}(t) = - \delta(t) v(t)
\end{cases}
\text{ and } \quad
\begin{cases}
\dot{u}(t) = v(t) \\
\dot{v}(t) = -  \nabla f(u(t)).
\end{cases}
\end{align*}
Now, using the numerical discretization \ref{Pb-1-discrete} on the first subproblem and \ref{Pb-2-discrete-II} on the second one, we obtain our second splitting algorithm. Now, we point out that, in our construction, we have used the fact that at $t=nh$, $\beta_n = (t-h)/(t+2h)$, where $h > 0 $ is the chosen step-size. This fact motivates the idea that we can take $\delta(t) = (t-\delta)/(t+2 \delta)$, such that it is the continuous counterpart of $\beta_n$. Furthermore, choosing $\delta > 0$ equal to $h$, we obtain that $\delta(nh) = \beta_n$, for each $n \in \mathbb{N}$. On the other hand, as $t \to \infty$, as in the case of the inertial sequence $(\beta_n)_{n \in \mathbb{N}}$, $\delta(t) \to 1$ as $t \to \infty$ and so, both the dynamical systems \ref{SSA1-DynSys} and \ref{SSA2-DynSys} are converging to \ref{SeqSplittI-DiffSyst} asymptotically. This is the first motivation why, in Subsection \ref{SSA1_SubSection} and in Subsection \ref{SSA2_SubSection} we have taken \ref{SeqSplittI-DiffSyst} to be a common ground to both our optimizers. The second motivation is that \ref{SeqSplittI-DiffSyst} is a linear autonomous evolution equation, so in this sense, it is much natural to apply a sequential splitting to this dynamical system. \\
Now, in Section \ref{Results_Section},  we have three subsections. In Subsection \ref{Algorithms_SubSection}, we present the basic concepts of machine learning algorithms. We deal with Adam, Adadelta, Adagrad, SGD, NaG and RMSProp in terms of mini-batches, epochs and iterations. The algorithmic terminology is deeply inspired by Ruder's work \cite{Ruder}. In Subsection \ref{DNN_SubSection}, we present the structure of the neural networks that are used in the training process on the following datasets: MNIST, MNIST-Fashion and CIFAR-10. In Subsection \ref{Experiments_SubSection}, we present our results regarding the comparison of our optimization algorithms with the classical ones belonging to the machine learning literature. Furthermore, we present the simulations concerning the loss, the accuracy and the training time of these optimizers. Finally, in Subsection \ref{Discussions_Section} we point out the novelty, the conclusions and the open problems related to both the theoretical and the computational aspects of our splitting-based numerical methods.
\color{black}

\section{Methods}\label{Methods_Section}

In this section, we present our proposed algorithms that drawn upon the operator splitting technique. The idea of symmetric Strang splitting has its roots back to \cite{Strang} and \cite{Marchuk}. The numerical methods that are based upon the idea of Strang splitting or sequential splitting (see \cite{Farago}  on Page 444) are known to be second order and first order, respectively. They are often used in solving ordinary and partial differential equations (see \cite{Holden}). Briefly, the operator splitting technique is a numerical method for the semi-discretization of a linear system of ODE's of the following abstract form

\begin{align*}
\dot{X}(t) = (A+B) X(t) \ ,
\end{align*}

where $A$ and $B$ are two given matrices. \\
The sequential splitting method is given by the approximation of two linear sub-problems, namely

\begin{align*}
\dot{X}(t) = A X(t) \text{ and } \dot{X}(t) = B X(t) \ .
\end{align*}

The full discretizations of these continuous sub-problems are combined, such that the iteration values for the first discretization is the previous value in the second discretization scheme. Normally, one has that 

\begin{align*}
e^{A+B} = e^A e^B \ ,
\end{align*}

if and only if $A$ and $B$ commute. At the same time, in the field of numerical analysis of evolution equations, the major interest lies in finding the theoretical order of convergence of the semi-discretization splitting method, that is buttressed by the evaluation of the absolute error at the first discretization point between the exact and the numerical splitting solution. In the case of Lie (sequential) splitting, this satisfies

\begin{align*}
\| e^{(A+B)h} - e^{Ah} e^{Bh} \| \leq \mathcal{O}(h) \ ,
\end{align*}

One the other hand, in the framework of optimization problems, one needs to focus the attention on the asymptotic convergence, on infinite intervals, since one is interested in the limit of the solution of the underlying dynamical system. Furthermore, in \cite{Hansen}, Hansen et. al. have proved the convergence of the Strang semi-discretization method in the setting of linear evolution equations. The case of semilinear evolution equations can be found in \cite{Ostermann}. \\
We observe that our algorithms start from a dynamical system that is similar to that of Nesterov's, i.e. \ref{NaG-DiffSyst}. But, our arguments are purely formal, in the sense that we adopt an intuitive way to infer a set of splitting-based numerical algorithms. We use some algebraic manipulations that will lead to some optimization schemes, using the idea of velocity given a second-order dynamical system. Firstly, we shall consider some remarks about the Nesterov's dynamical system \ref{NaG-DiffSyst}, that will be crucial for the constant-damping dynamical systems of our optimization schemes.

\subsection{Sequential splitting algorithm I}\label{SSA1_SubSection}

We consider the following dynamical system with constant damping:
\begin{align}\label{SeqSplittI-DiffSyst}
\ddot{u}(t) + \dot{u}(t) = -\nabla f(u(t)).
\end{align}
In the vein of Nesterov's algorithm \ref{NaG}, one can easily apply a forward Euler or a Crank-Nicolson type method to \ref{SeqSplittI-DiffSyst}, such that it preserves the properties of the continuous dynamical system. Our approach is rather empirical: we split the continuous dynamical system into two sub-problems and discretize each of them with some basic numerical methods. After that, we merge these two discretization into one algorithm. This numerical procedure will lead to different velocity updates for the current iteration value. \\
Now, the first subproblem will contain the linear part of the velocity updates and is the most basic one:

\begin{align}\label{Pb-1}
\begin{cases}
\dot{u}(t) = 0 \\
\dot{v}(t) = - v(t).
\end{cases}
\end{align}

Further, the second subproblem will contain the continuous gradient of the objective function $f$, as follows:

\begin{align}\label{Pb-2}
\begin{cases}
\dot{u}(t) = v(t) \\
\dot{v}(t) = - \nabla f(u(t)).
\end{cases}
\end{align}

On the other hand, for the subproblem \ref{Pb-1}, we employ the explicit forward Euler discretization, as follows 

\begin{align}\label{Pb-1-discrete}
\begin{cases}
u_{n+1/2} = u_{n} \\
\dfrac{v_{n+1/2}-v_n}{h} = - \beta_{n} v_{n},
\end{cases}
\end{align}

where $u_{n+1/2}$ is the iteration term for the discretization of the continuous differential system \ref{Pb-1}. Further, we know that $\beta_n$ converges to $1$ as $n$ goes to $+\infty$, so \ref{Pb-1-discrete} is a proper numerical discretization, as time grows to infinity. \\
At the same time, for the second continuous subproblem \ref{Pb-2}, we consider the symplectic Euler numerical algorithm for the iteration value $u_n$ and for the velocity updates $v_n$, in the following way:

\begin{align}\label{Pb-2-discrete-original-I}
\begin{cases}
\dfrac{u_{n+1} - u_{n+1/2}}{h} = \beta_n^2 v_{n+1} \\
\dfrac{v_{n+1} - v_{n+1/2}}{h} = - \nabla f(y_n).
\end{cases}
\end{align}

Now, we introduce $\hat{v}_{n+1}$ which is a perturbed velocity under the value of the gradient. That is

$$\hat{v}_{n+1} = v_{n+1} - h \cdot \left( \dfrac{1}{\beta_n^2}-1 \right) \nabla f(y_n).$$
So, we modify our algorithm on the second sub-problem \ref{Pb-2} in the following manner:

\begin{align}\label{Pb-2-discrete-I}
\begin{cases}
\dfrac{u_{n+1} - u_{n+1/2}}{h} = \beta_n^2 \hat{v}_{n+1} \\
\dfrac{v_{n+1} - v_{n+1/2}}{h} = - \nabla f(y_n).
\end{cases}
\end{align}

If we combine \ref{Pb-2-discrete-I} with \ref{Pb-1-discrete}, after some algebraic manipulations, we will eventually obtain the following algorithm:

\begin{align}\label{SeqSplitt-I}
\begin{cases}
u_{n+1} = u_{n} + h \beta_n^2 (1-h \beta_n) v_n  - h^2 \nabla f(y_n) \\
v_{n+1} = (1-h \beta_n) v_n - h \nabla f(y_n),
\end{cases}
\end{align}

where the additional iteration $y_n$ will depend on $u_n$ and $v_n$. Now, the most intuitive approach for our discretization is to take $\hat{y}_n = u_n + h \beta_n^2 (1-h \beta_n) v_n$, in order to have $u_{n+1} = \hat{y}_n - h^2 \nabla f(y_n)$, which is similar to the case of Nesterov's algorithm \ref{NaG}. From the preliminary results of the numerical simulations we infer that a better approach would be to take the same value for the momentum iteration updates $y_n$ as in the case of \ref{NaG}, i.e. 

\begin{align}\label{MomIterate}
y_n = u_n + h \beta_n v_n \ .
\end{align}

Finally, we consider the modified form of the algorithm \ref{SeqSplitt-I}, using the addition of a hyper-parameter $k$ that will boost the velocity values in order to reach the minimum of the objective function at least as fast as \ref{NaG}:

\begin{align}\label{SeqSplitt-I-Alg}
\begin{cases}
y_n = u_n + h \beta_n v_n \\
v_{n+1} =  \beta_{n}^{k} \cdot ((1-h \beta_n) v_n - h \nabla f(y_n)) \\
u_{n+1} = u_{n} + \beta_n (1-h \beta_n) (y_n-u_n)  - h^2 \nabla f(y_n),
\end{cases}
\end{align}

where the hyper-parameter $k$ is chosen such as $k \geq 0$.

\subsection{Sequential splitting algorithm II}\label{SSA2_SubSection}

In the case of the second algorithm we propose, we also employ the sequential splitting. We consider the dynamical system \ref{SeqSplittI-DiffSyst}, which we divide into the same two sub-problems as before, that is \ref{Pb-1} and \ref{Pb-2}. For the first sub-problem, we employ the same discretization, namely the forward Euler method \ref{Pb-1-discrete}. On the other hand, for the second sub-problem we consider the following discrete counterpart, namely also the explicit Euler method:

\begin{align}\label{Pb-2-discrete-II}
\begin{cases}
\dfrac{u_{n+1} - u_{n+1/2}}{h} = v_{n+1/2} \\
\dfrac{v_{n+1} - v_{n+1/2}}{h} = - \nabla f(y_n).
\end{cases}
\end{align}

In this case, we observe that for the principal iterate values $u_{n+1}$, we have used the previous value updates of the velocity, namely $v_{n+1/2}$. In this sense, our sequential splitting algorithm is of backward type. Merging \ref{Pb-1-discrete} and \ref{Pb-2-discrete-II}, we obtain the following optimization algorithm that is based upon the sequential splitting technique:

\begin{align}\label{SeqSplitt-II}
\begin{cases}
u_{n+1} = u_{n} + h (1-h \beta_n) v_n\\
v_{n+1} = (1-h \beta_n) v_n - h \nabla f(y_n),
\end{cases}
\end{align}

where, as before, $y_n$ will depend on $u_n$ and on the velocity values $v_n$. One can notice that the natural choice for $y_n$ is $u_{n+1}$. As in \ref{SeqSplitt-I}, we shall opt for $y_n = u_n + h \beta_n v_n$. Subsequently, one obtains that the value of $h v_n$ is equal to $ (y_n-u_n) \cdot \dfrac{1}{\beta_n}$. \\
Taking these into account, we consider the following inertial algorithm that will be appropriate for unconstrained optimization problems:

\begin{align*}
\begin{cases}
y_n = u_n + h \beta_n v_n \\
v_{n+1} = (1-h \beta_n) v_n - h \nabla f(y_n) \\
u_{n+1} = u_{n} + \dfrac{1-h\beta_n}{\beta_n} (y_n-u_n),
\end{cases}
\end{align*}

Also, in order to improve the empirical rate of convergence of our algorithm, we will add one hyper-parameter $k \geq 0$. From our numerical computations we set the default value of $k$ equal to $2.0$. So our proposed algorithm is the following:

\begin{align}\label{SeqSplitt-II-Alg}
\begin{cases}
y_n = u_n + h \beta_n v_n \\
v_{n+1} = \beta_{n}^{k} \cdot ((1-h\beta_n) v_n - h \nabla f(y_n)) \\
u_{n+1} = u_{n} + \dfrac{1-h\beta_n}{\beta_n} (y_n-u_n).
\end{cases}
\end{align}

Finally, we observe that our new algorithm \ref{SeqSplitt-II-Alg} differs from the first splitting algorithm \ref{SeqSplitt-I-Alg} in the iterates values $u_{n+1}$, since the latter algorithm does not contain the gradient of the objective function $f$. In addition to this, the velocity values $v_n$ are updated with different parameters. In the framework of neural network training they depend on the current iteration, where the gradient is computed only on a mini-batch dataset.

\begin{remark}{\ \\}
If we compare our optimization schemes \ref{SeqSplitt-I-Alg} and \ref{SeqSplitt-II-Alg}, we can see that the latter one is a proper discretization scheme of the associated  asymptotic  dynamical system. Further, the first splitting based algorithm has a different  asymptotic  dynamical system, by the fact that the velocity used in the second sub-problem \ref{Pb-2} is perturbed with the value of the gradient in the point $u_n + h \beta_n v_n$.
\end{remark}

\section{Results}\label{Results_Section}

\subsection{ A gentle introduction to machine learning algorithms}\label{Algorithms_SubSection}

In this section, we shall review some optimization algorithms that are often used in the training of deep and convolution neural networks. In our neural network training, we use only the stochastic variant of these algorithms, including epoch-training with mini-batches. Furthermore, we shall present algorithmically our optimization schemes \ref{SeqSplitt-I-Alg} and \ref{SeqSplitt-II-Alg}, that will be also used with their stochastic counterpart. From now on, the first splitting-based algorithm \ref{SeqSplitt-I-Alg} will be called \color{red} \emph{SSA1} \color{black}. \\
In the neural network training, the objective (also called loss function) must be minimized with respect to some parameters. In general, these parameters contain the weights and biases in the neural network. From now on, we consider $\theta$ to be the collection of weights and biases and $(x^{(i)}, y^{(i)})$ to be the given training input values and target values, respectively. \\
In the first section, we have introduced the most basic optimization algorithm \ref{GD}, namely gradient descent scheme.
More often it is used the stochastic variant of this algorithm, namely stochastic gradient descent (see \cite{Bottou}  on Page 2 ). In the full version of this algorithm, namely mini-batch stochastic gradient descent, one updates the parameter iteration values for each shuffled mini-batch $m$ of training examples is given as follows

\begin{align}\label{MiniBatch-SGD}
\theta_{n+1} = \theta_{n} - h \nabla f_{\theta}(\theta_n ; x^{(i:i+m)}, y^{(i:i+m)}).
\end{align}

This algorithm leads to a more stable convergence than plain \ref{GD}, but updates the parameters with high variance (see \cite{Ruder}) and will eventually lead to oscillations in the decrease of the objective function. The advantage is that at each epoch we dispose of redundant data input values, since at each iteration we compute the gradient on a mini-batch of a given size $m$. Despite the clear disadvantage of using the same learning rate (step-size) $h$ for every component of the vector $\theta_{n+1}$, the algorithm has lower chances than plain \ref{GD} to be stuck at a saddle point of the objective function. \\
Other iteration schemes that are deployed in neural network training are the adaptive algorithms. The most basic one is Adagrad (see \cite{Adagrad} on Page 2122). This is used for large-scale neural networks, but has the disadvantage that it has a very aggressive learning rate update. So, a modification of this algorithm is Adadelta (see \cite{Adadelta}  on Page 3), in which one stores previous values of the squared gradient of the objective function. Further, it is known that decreasing the effect of the past gradients lead to less sensitivity in choosing the hyper-parameters of the neural model. In Adadelta, the so-called running average is defined as 

\begin{align*}
E[\nabla^2 f_{\theta}(\theta_{n} ; x^{(i:i+m)}, y^{(i:i+m)})] = \gamma \cdot E[\nabla^2 f_{\theta}(\theta_{n-1} ; x^{(i:i+m)}, y^{(i:i+m)})] + (1-\gamma) \cdot \nabla^2 f_{\theta}(\theta_n ; x^{(i:i+m)}, y^{(i:i+m)}),
\end{align*}

where $\gamma$ is taken, in general, 0.9. Also, for each epoch and for a chosen mini-batch $(x^{(i:i+m)}, y^{(i:i+m)})$ of a given suitable size $m$, another exponential decay average is defined, such that

\begin{align*}
E[\Delta \theta^2]_n = \gamma \cdot E[\Delta \theta^2]_{n-1} + (1-\gamma) \cdot \Delta \theta^2_n,
\end{align*}

with the remark that the update values are given by the negative ratio of two adaptive rates. The coefficient from the nominator contains the square root of the exponential energy decay and the denominator is the square root of the past gradients. For more details, we let the reader follow \cite{Adadelta}. \\
Another popular algorithm in neural network training is RMSprop which was introduced in the Coursera lecture class by G. Hinton \cite{RMSProp}  on Slide 26. Following \cite{Ruder}, RMSprop algorithm is in fact the same as Adadelta's vector update, using the adaptive learning rate of the form

\begin{align*}
h_{RMSprop} = - \dfrac{h}{\sqrt{E[\nabla^2 f_{\theta}(\theta_{n} ; x^{(i:i+m)}, y^{(i:i+m)})]+\varepsilon}} \ .
\end{align*}

In general, the default value for the starting learning rate is of order of magnitude $10^{-3}$. One can observe that in the case of RMSprop algorithm, the effective value for the adaptive learning rate depends on the magnitude of the squared energy of the past gradient values, for each vector component of the underlying parameter. \\
Last but not least, we recall the optimization algorithm Adam (see \cite{Adam}  on Page 2), that can be considered as a modified gradient descent using two biased corrected moments, i.e.

\begin{align*}
\begin{cases}
mom_n = \beta_1 mom_{n-1} + (1-\beta_1) \nabla f_{\theta}(\theta_{n} ; x^{(i:i+m)} , y^{(i:i+m)}) \\
v_n = \beta_2 v_{n-1} + (1-\beta_2) \nabla^2 f_{\theta}(\theta_{n} ; x^{(i:i+m)}, y^{(i:i+m)}) \\
\hat{mom}_n = \dfrac{mom_n}{1-\beta^n_1} \\
\hat{v}_n = \dfrac{v_n}{1-\beta^n_2} \\
\theta_{n+1} = \theta_n - \dfrac{h}{\sqrt{\hat{v}_n}+\varepsilon} \cdot \hat{mom}_n,
\end{cases}
\end{align*}

where, as before, $m$ is the size of the mini-batch used in the current iteration $n$. So, Adam is a modified adaptive gradient descent type-scheme using two moments that are corrected in each iteration.  Also, in the machine learning community, Adam seems to be the most popular optimization algorithm and one of the most easy optimizers that require minimal tuning of the parameters (see \cite{Karpathy}).  From \cite{Nadam}  Page 3, we recall the Nadam algorithm, which basically represents the combination between Adam and the momentum method. In \cite{Sutskever}, it is mentioned that this algorithm is comparable to first order adaptive methods. The inertial parameter in the momentum method leads to higher quality updates of the weights and biases values, where in one updates the parameters after the momentum correction of the past gradients. Last but not least, we reassert that instead of using adaptive learning rate algorithms, one can use \ref{MiniBatch-SGD} using learning rate updates that depend on the current epoch training(for more details, follow \cite{CyclicLearningRate}). For some references regarding the choice of the optimal learning rate, the optimizers for neural networks and for an empirical analysis of the training and classification problems, we refer to \cite{Bengio}, \cite{BottouNocedal}, \cite{Zhang} and \cite{Bishop}. \\
In classification-type tasks, stochastic gradient descent \ref{MiniBatch-SGD} is better the generalization errors obtained on the validation and test datasets than adaptive learning rate algorithms. For a full discussion of this problem, we refer to \cite{Wilson}. Also, for some general notions concerning deep learning principles, classification and optimization problems we refer to \cite{Nielsen}, \cite{ML_Physics}, \cite{Higham}, \cite{LeCun} and \cite{Goodfellow}. Now, at the end of this section we consider the stochastic version of the \color{red} \emph{SSA1} \color{black} optimizer, that is \ref{SeqSplitt-I-Alg}. This is presented below, as Algorithm \ref{alg:SSA1}. Moreover, in our algorithms the current iteration is denoted by $n$ and the learning rate by $h$.

\begin{algorithm}
\caption{Mini-Batch Stochastic SSA1}\label{alg:SSA1}
\begin{algorithmic}[1]
    \State \textbf{Require: } Learning rate $h$
    \State \textbf{Require: } Gradient Information $\nabla$ for the data $f$
    \State \textbf{Require: } Initial parameter $\theta_{0}$
    \State \textbf{Require: } Hyper-parameter $k$
    \For {i \textbf{in range} (epochs)}
    \State \textbf{Shuffle data input values} 
    \For {batchSample $(x^{(i:i+m)}, y^{(i:i+m)})$}
    \State \color{blue} Update Inertial Parameter\color{black}: $\beta_n = \dfrac{n}{n+3}$
    \State \color{blue} Update Information\color{black}: $\theta_{n+1} = \theta_{n} + h (1-h \beta_n) v_n  - h^2 \nabla f_{\theta}(\theta_n + h \beta_n v_n ; x^{(i:i+m)}, y^{(i:i+m)})$
    \State \color{blue} Update Velocity\color{black}: $v_{n+1} = \beta_{n}^{k} \cdot \left[ (1-h \beta_n) v_n - h \nabla f_{\theta}(\theta_n + h \beta_n v_n ; x^{(i:i+m)}, y^{(i:i+m)}) \right]$
	\EndFor
    \EndFor
\end{algorithmic}
\end{algorithm}

\begin{algorithm}
\caption{Mini-Batch Stochastic SSA2}\label{alg:SSA2}
\begin{algorithmic}[1]
    \State \textbf{Require: } Learning rate $h$
    \State \textbf{Require: } Gradient Information $\nabla$ for the data $f$
    \State \textbf{Require: } Initial parameter $\theta_{0}$
    \State \textbf{Require: } Hyper-parameter $k$
    \For {i \textbf{in range} (epochs)}
    \State \textbf{Shuffle data input values} 
    \For {batchSample $(x^{(i:i+m)}, y^{(i:i+m)})$}
    \State \color{blue} Update Inertial Parameter\color{black}: $\beta_n = \dfrac{n}{n+3}$
    \State \color{blue} Update Information\color{black}: $\theta_{n+1} = \theta_{n} + \dfrac{h (1-h \beta_n) \beta_n}{\beta_n} v_n$
    \State \color{blue} Update Velocity\color{black}: $v_{n+1} = \beta_{n}^{k} \cdot \left[ (1-h \beta_n) v_n - h \nabla f_{\theta}(\theta_n + h \beta_n v_n ; x^{(i:i+m)}, y^{(i:i+m)}) \right]$
	\EndFor
    \EndFor
\end{algorithmic}
\end{algorithm}

At the end of this sub-section, we present the mini-batch stochastic algorithms that are similar to the adaptive Adadelta optimizer, namely Algorithm \ref{alg:SSA1-Ada}. In the numerical computations, we shall briefly call it \color{red} SSA1-Ada \color{black}.

\begin{algorithm}
\caption{\emph{Adaptive} Mini-Batch Stochastic SSA 1}\label{alg:SSA1-Ada}
\begin{algorithmic}[1]
\State \textbf{Require: } Learning rate $h$
\State \textbf{Require: } Decay rate $\rho$
\State \textbf{Require: } Term preventing division by zero $\varepsilon$
\State \textbf{Require: } Gradient Information $\nabla$ for the data $f$
\State \textbf{Require: } Initial parameter $\theta_0$
\State \textbf{Require: } Hyper-parameter $k$
\For {i \textbf{in range} (epochs)}
    \State \textbf{Shuffle data input values} 
    \For {batchSample $(x^{(i:i+m)}, y^{(i:i+m)})$}
    \State \color{blue} Update Inertial Parameter\color{black}: $\beta_n = \dfrac{n}{n+3}$
    \State \color{blue} Accumulate Gradient\color{black}: $E[\nabla^2 f_{\theta}(z_{n} ; x^{(i:i+m)}, y^{(i:i+m)})] = \rho \cdot E[\nabla^2 f_{\theta}(z_{n-1} ; x^{(i:i+m)}, y^{(i:i+m)})] + $ \\
    \hspace*{10cm} $(1-\rho) \cdot \nabla^2 f_{\theta}(z_n ; x^{(i:i+m)}, y^{(i:i+m)})$
    \State \color{blue} Root Mean Square of $f$\color{black}: $RMS[\nabla f]_n = \sqrt{E[\nabla^2 f_{\theta}(z_{n} ; x^{(i:i+m)}, y^{(i:i+m)})] + \varepsilon}$
    \State \color{blue} Compute Adaptive step-size\color{black}: $h_n = h \cdot \dfrac{RMS[\Delta z]_{n-1}}{RMS[\nabla f]_n}$
    \State \color{blue} Compute Updates\color{black}: $\Delta z_{n} = - h_n \cdot \nabla f_{\theta}(z_n ; x^{(i:i+m)}, y^{(i:i+m)})$
    \State \color{blue} Accumulate Updates\color{black}: $E[\Delta z^2]_{n} = \rho \cdot E[\Delta z^2]_{n-1} + (1-\rho) \cdot \Delta z^2_n$
    \State \color{blue} Root Mean Square of the update\color{black}: $RMS[\Delta z]_{n} = \sqrt{E[\Delta^2 z]_{n} + \varepsilon}$
    \State \color{blue} Compute Additional Iteration\color{black}: $z_{n+1} = \theta_n + h \beta_n v_n$
    \State \color{blue} Update Velocity\color{black}: $v_{n+1} = \beta_n^k \cdot \left[ (1-h_n \beta_n) v_n - h_n \nabla f_{\theta}(z_{n+1} ; x^{(i:i+m)}, y^{(i:i+m)}) \right]$
    \State \color{blue} Update Information\color{black}: $\theta_{n+1} = \theta_n + \beta_n (1 - h_n \beta_n
    ) \cdot (z_{n+1} - \theta_n) - h_n^2 \nabla f_{\theta}(z_{n+1} ; x^{(i:i+m)}, y^{(i:i+m)})$
\EndFor
	\EndFor    
\end{algorithmic}
\end{algorithm}

\begin{remark}{\ \\}
From our numerical results we have observed that the adaptive counterpart of \color{red} SSA 1 \color{black} algorithm, namely \color{red} SSA 1-Ada \color{black} is competitive with all other adaptive-type schemes. A similar adaptive algorithm can be given for \color{red} SSA 2 \color{black}. Further, also from our computations we have observed that this algorithm reaches almost $10\%$ accuracy on both MNIST and MNIST-Fashion. So, our results from the last sub-section does not include the adaptive mini-batch variant of our second proposed optimizer.  
\end{remark}

\subsection{Description of the neural network}\label{DNN_SubSection}

For our numerical simulations we have used some adaptive and non-adaptive stochastic algorithms with mini-batches in order to minimize the loss function of a convolution neural network. We have used both well known Hand Written digit Recognition Dataset (MNIST) from \url{http://yann.lecun.com/exdb/mnist/} and the harder drop-in replacement MNIST-Fashion presented in \url{https://github.com/zalandoresearch/fashion-mnist}. For the neural network we have used the MNIST example provided by \texttt{Pytorch} at \url{https://github.com/pytorch/examples/tree/master/mnist}. The CNN has a convolution layer with a kernel size of 5 pixels which converts the single channel input to a 20 channel output. After RELU is applied to the output of this layer, it is max pooled with a kernel of 2 pixels and is passed as the input to another convolution layer wit the same kernel size, but whose number of output channels is 50. RELU and the same max pooling technique are applied again. This output is flattened and then is connected to a layer of 500 hidden neurons. The output of this hidden layer is passed through another RELU and then is connected to a 10 neuron output layer. The final output is obtained via the \emph{logsoftmax} activation function of the output layer activation values, given by the formula
\begin{align*}
        f(\bf{x}^{[i]}) = \log \left( {\frac{\exp(\bf{x}^{[i]})}{\sum\limits_{i=0}^{n} \exp(\bf{x}^{[i]})}} \right) ,
\end{align*}

where $\bf{x}^{[i]}$ represent the $i^{th}$ component of the entry vector $\bf{x}$. \\
Also, this model does not use any dropout and for the loss function we have used the \emph{Negative Log Likelihood} given by the formula

\begin{align*}
        l(\bf{x}, \bf{y}) = -\log \left( {\bf{x}^{i}} \right) \, \textbf{ if } \bf{y}^{i} = 1.    
\end{align*}

Furthermore, we have adapted the code provided by \texttt{Pytorch}'s example in order to facilitate the choice of different optimizing algorithms and different loss functions. The new algorithms we introduced were implemented as classes in the \texttt{Pytorch} \texttt{optim} package, using a coding style as similar as possible to the style used by the \texttt{Pytorch} developers. For the experiments we have employed the standard split for both MNIST and MNIST-Fashion, so the 60000 images were divided in two sets, one with 50000 images for training and one with 10000 images for testing. We did not use validation data because we desired to mimic the results obtained by the example provided by \texttt{Pytorch}. The data sets used were loaded via the \texttt{Pytorch} loaders and were normalized with mean $0.1307$ and standard deviation $0.3081$. We divided both the training and test data in mini-batches and then we did our training in a stochastic manner. For the shuffling of the mini-batches we used the random seed 1. We have also used the seed as \texttt{Pytorch}'s seed for the random initialization of the weights and biases. We saved the model that we started the experiments using \texttt{Pytorch}'s built-in save feature. \\
On the other hand, CIFAR 10 (\url{https://www.cs.toronto.edu/~kriz/cifar.html}) is a dataset of images from 10 different categories. There are 60000 color square images measuring 32 by 32 pixels. The dataset was proposed by the Canadian Institute for Advanced Research. In our computations, we have used the \texttt{GoogLeNet} model proposed in \cite{Szegedy} by Szegedy et. al. The model contains 22 inception layers and has achieved an error rate of 6.67\% in the top 5 categories. We chose this model because it is circumscribed by the 1.5 billion add-multiply operations budget, which in turn means a faster, shorter training time for our optimizer.
The model's implementation is available on GitHub at \url{https://github.com/kuangliu/pytorch-cifar.git}. In order to load the data, we have used the methods provided by \texttt{Pytorch} datasets package. On top of this, we emphasize the fact that our results on \texttt{CIFAR 10} lie upon the use of the \emph{Cross Entropy Loss}, which is a combination between \textit{Negative Log Likelihood} loss and \textit{softmax} activation function. \\
Moreover, for the simulation that entails \texttt{CIFAR 10}, we employed the following learning rates: for the adaptive algorithms Adadelta, Algorithm \ref{alg:SSA1-Ada}, we took the default learning rate 1.0. Also, for the other optimizers, i.e. RMSProp, Adagrad , Adam , \ref{MiniBatch-SGD}, \ref{NaG}, Algorithm \ref{alg:SSA1} and Algorithm \ref{alg:SSA2}, we took the learning rate 0.001, since this represents a good value for 200 epochs of training. \\
Last but not least, we emphasize the fact that our numerical computations were made on a \texttt{GPU:NVIDIA Tesla V100-SXM2}, which is a GPU data center that has 5120 NVIDIA Cuda cores, GPU memory 16 GB HBM2 and its double precision performance is around 7.5 TFlops. In a nutshell, we point out that in \texttt{Keras} and \texttt{Tensorflow}, the stochastic algorithms with mini-batches present heavy oscillations in the decrease of the loss function, since the graphical representations of the loss function are made with respect to each iteration concerning mini-batches. In our case, the plots are given in terms of accuracy and loss values at the end of each epoch, and not in terms of each particular iteration at different mini-batch datasets.

\subsection{ The neural network experiments}\label{Experiments_SubSection}

\begin{figure}[!ht]
     \centering
     \begin{subfigure}[b]{0.49\textwidth}
         \centering
         \includegraphics[width=\textwidth]{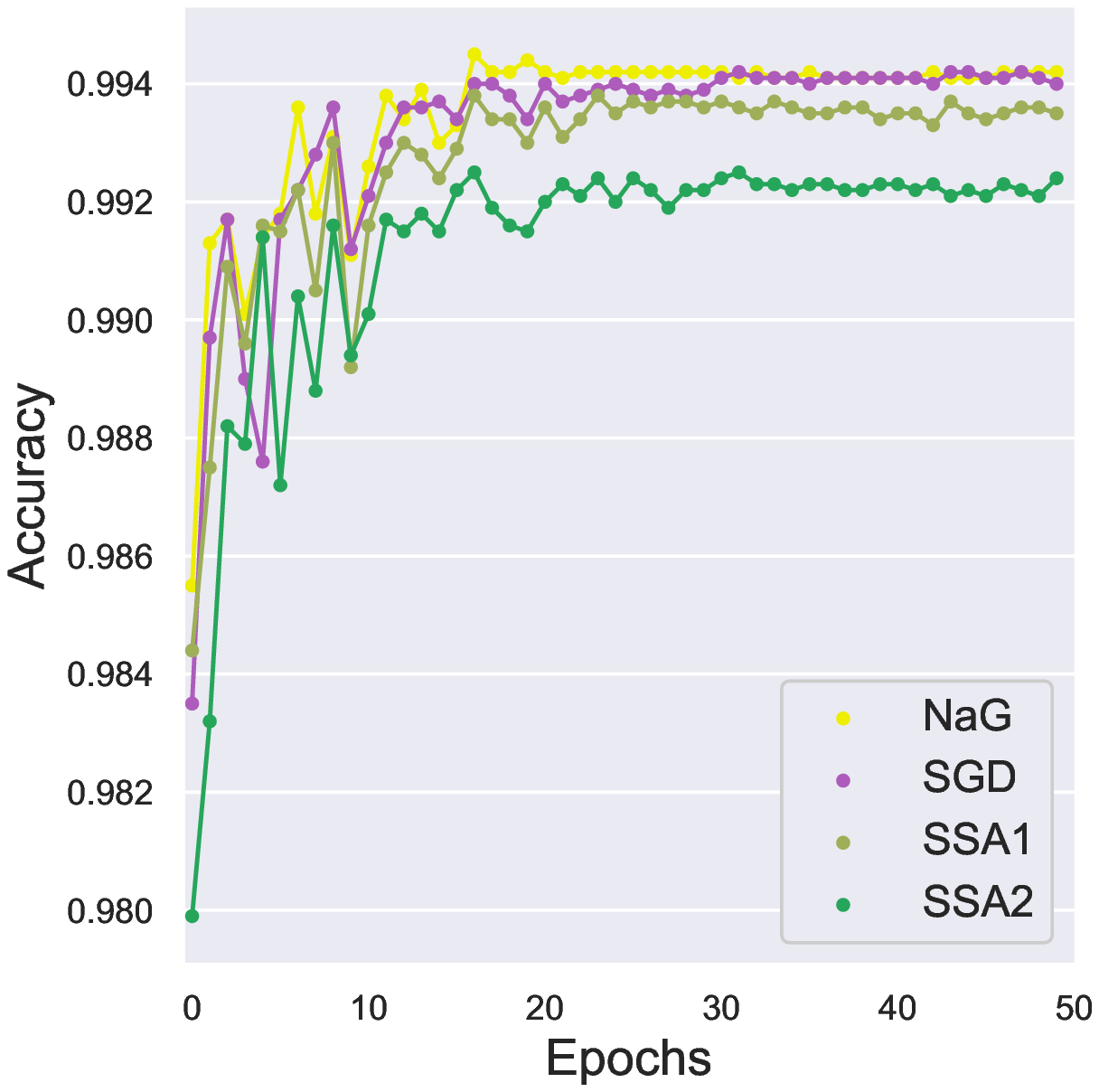}
         \caption{Accuracy comparison.}
         \label{fig:ComparisonMNIST_Acc}
     \end{subfigure}
     \hfill
     \begin{subfigure}[b]{0.49\textwidth}
         \centering
         \includegraphics[width=\textwidth]{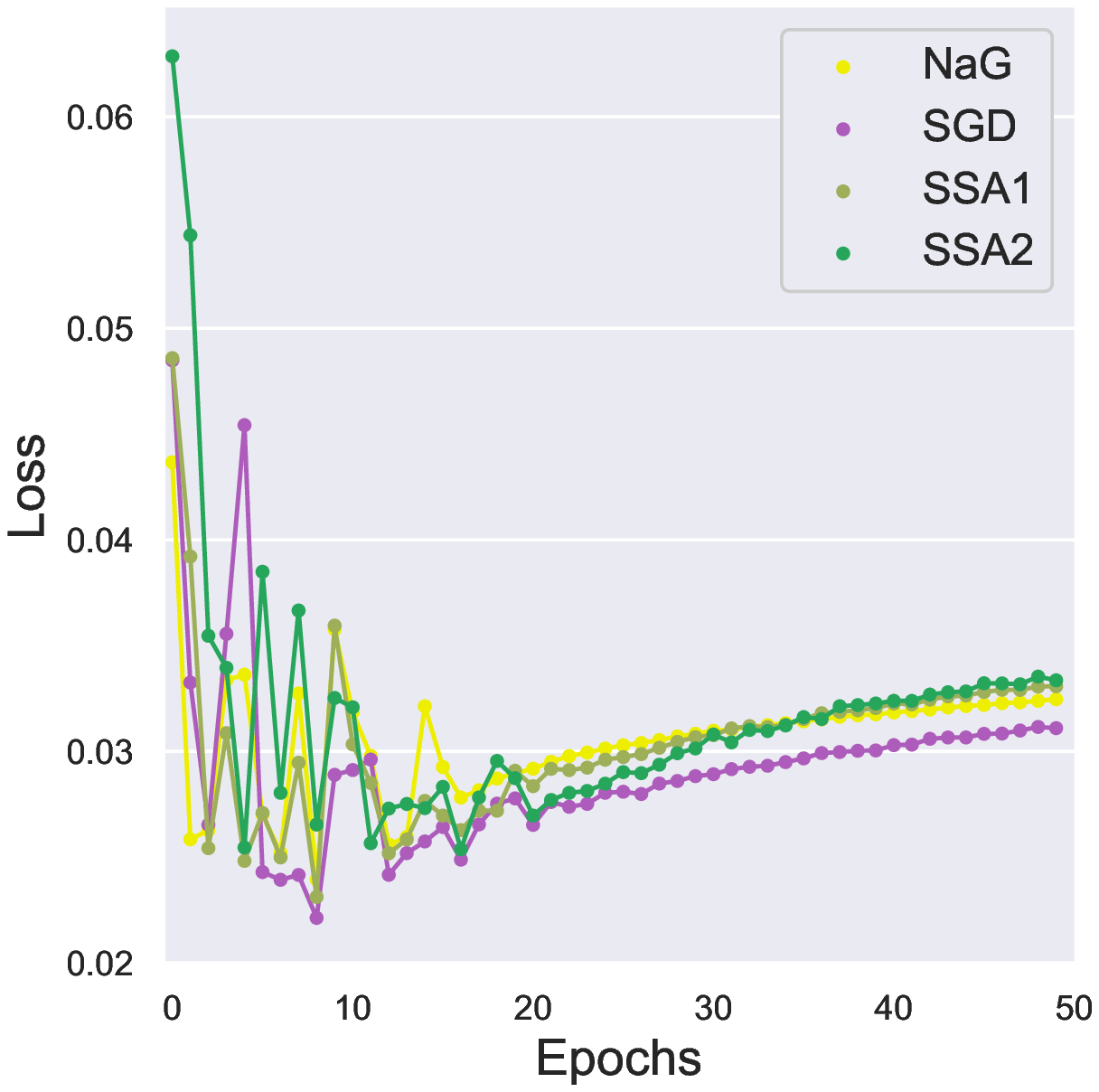}
         \caption{Comparison for the decrease in the loss function.}
         \label{fig:ComparisonMNIST_Loss}
     \end{subfigure}
        \caption{Comparison on the test data for the \ref{MiniBatch-SGD}, \ref{NaG}, Algorithm \ref{alg:SSA1} and Algorithm \ref{alg:SSA2}, on the basic MNIST dataset. Number of epochs 50, learning rate 0.1.}
        \label{fig:MNIST_figures}
\end{figure}

In Figure \ref{fig:ComparisonMNIST_Acc}, we plotted the overall accuracy on the MNIST test dataset. The learning rate of \ref{MiniBatch-SGD}, \ref{NaG}, Algorithm \ref{alg:SSA1} and Algorithm \ref{alg:SSA2} has been set to a high value, i.e. $0.1$ and the graph represents the accuracy values on 50 epochs. Our algorithms are comparable to the stochastic version of the stochastic gradient descent and with the Nesterov's algorithm. On this graph we observe that our optimizers have a better accuracy on the MNIST dataset, but present higher oscillations due to their velocity updates that are based on the constant parameters $k$ and $q$. \\
Furthermore, since the decrease in the loss function presents the same behavior as in the increase of the accuracy from Figure \ref{fig:ComparisonMNIST_Loss}, we have presented the decrease in the objective function. We have stuck to the same 50 epochs an  d 0.1 learning rate of the chosen algorithms. We can observe that the loss function of the \ref{MiniBatch-SGD} showcases high oscillations over the first 20 epochs. Due to the inherent inertial moment, \ref{NaG} alleviates these oscillations and stabilizes the decrease of the objective function. Conversely, the loss value of these algorithms increases after 20 epochs and this can be prevented with some early-stopping techniques. Last but not least, \color{red} SSA 1 \color{black} and \color{red} SSA 2 \color{black} present a similar value with respect to the loss function as \ref{NaG}, on the test dataset, and this is in correlation with the increased accuracy observed in the Figure \ref{fig:ComparisonMNIST_Acc}.

\begin{figure}[!ht]
\hspace{-0.5cm} \includegraphics[width=15.2cm, height=7.5cm]{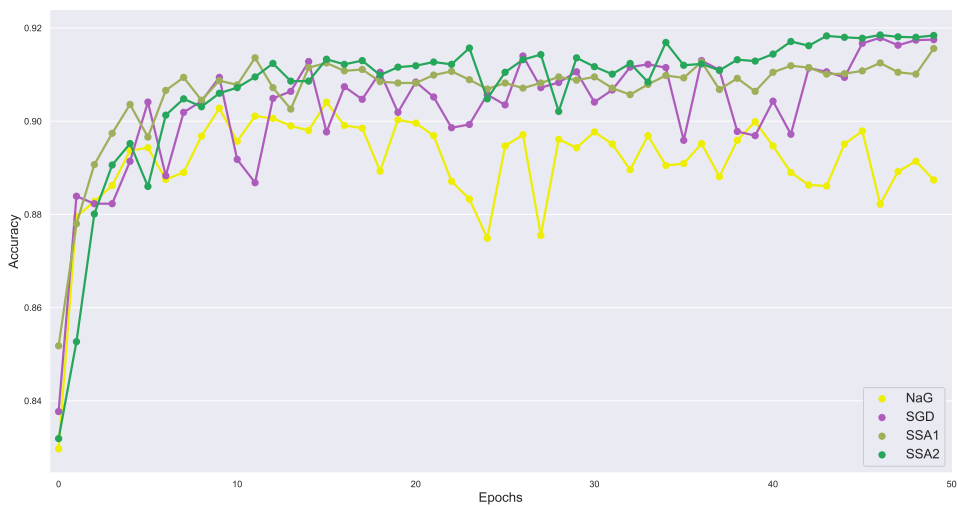}
\caption{Accuracy comparison on the test data for the \ref{MiniBatch-SGD}, \ref{NaG}, Algorithm \ref{alg:SSA1} and Algorithm \ref{alg:SSA2}, on the MNIST-Fashion dataset. Number of epochs 50, learning rate 0.1 }\label{fig:ComparisonFashMNIST_Acc}
\end{figure}

In Figure \ref{fig:ComparisonFashMNIST_Acc}, we considered the graph of the accuracy values of the same algorithms as before, but for the MNIST-Fashion test dataset. It is known that on the MNIST-Fashion dataset, the accuracy is not as high as in the classical MNIST dataset. Quite surprisingly, our newly introduced \color{red} SSA 1 \color{black} and \color{red} SSA 2 \color{black} seem to behave better than what we have seen in Figure \ref{fig:ComparisonMNIST_Acc}. Due to their velocity updates in the gradient values, especially in the sequential splitting Algorithm \ref{alg:SSA1}, these optimizers do not oscillate with respect to the increase in epochs. While this stands true for our splitting-based schemes, \ref{MiniBatch-SGD} and \ref{NaG} present powerful degradation in the accuracy on the same test dataset. Also, a similar analysis can be made for the decrease in the loss function, as in Figure \ref{fig:ComparisonMNIST_Loss}. \\
In Figures \ref{fig:ComparisonMNIST_Acc}, \ref{fig:ComparisonMNIST_Loss} and \ref{fig:ComparisonFashMNIST_Acc}, we have the behavior of our stochastic non-adaptive algorithms in the presence of the training of the convolution neural network. Further, \color{red} SSA 1 \color{black} and \color{red} SSA 2 \color{black} optimizers have some additional hyper-parameter $k$. This is clearly an advantage in the velocity updates of these discretizations. On the other hand, for more challenging classification problems, one needs to choose the optimal values for these parameters. Even though it is of good practice to employ in the numerical simulations a Bayesian optimization technique, it is easily verifiable that it is time consuming for these two parameters, in addition with the learning rate values. Furthermore, on more complex classification problems we have observed, through our numerical simulations, that moderate values of both $k$ , i.e. $k=2.0$ can be considered as default values for this optimization schemes, yielding to a good convergence behavior of our proposed algorithms.

So far we have put an emphasis on our numerical results on some basic experiments. All of our computations are presented in Table \ref{tab:MNIST_Results} for both MNIST and MNIST-Fashion datasets. Also, since the phenomenon of overfitting is crucial in machine learning, we have illustrated the accuracy on the training set (lower diagonal) and the accuracy on the test set (upper diagonal). For our non-adaptive optimizers we have set the step-size to vary from $0.001$ to $0.1$ and for our adaptive counterparts we have set the learning rate to the default value of $1$, since \color{red} SSA 1-Ada \color{black} is a combination between Adadelta, Algorithm \ref{alg:SSA1}. We focus our attention on a few remarks regarding our results for the accuracy values on the MNIST dataset. Both stochastic gradient descent and Nesterov's algorithm with momentum 0.5 have almost 0.99 accuracy on the test set, when the learning rate is equal to 0.1. Furthermore, \color{red} SSA 1 \color{black} and \color{red} SSA 2 \color{black} display almost the same behavior when the step-size is large enough, i.e. they have almost 0.992 and 0.993 accuracy value on the test set. On the other hand, a lower learning rate of approximate value of 0.001 leads to a difference in the train and test accuracy for the splitting algorithms, yet for \ref{MiniBatch-SGD} and \ref{NaG} it does not lead to the overfitting problem when lower step-size values are used. In addition, it seems that \ref{NaG} achieves better accuracy than both SGD and \color{red} SSA 1 \color{black} after 20 epochs until this value is stabilized. For \color{red} SSA 1 \color{black} and \color{red} SSA 2 \color{black} there seems to be a negligible difference in training versus testing accuracy after the first 50 epochs.
Now, regarding our adaptive algorithms, for the MNIST dataset, the Adadelta algorithm achieves a very good accuracy before 20 epochs of learning. The same stands true for \color{red} SSA 1-Ada \color{black}, but these algorithms degrade in their accuracy and the need for early-stopping is a serious issue. Only 20 epochs are needed in order to train our adaptive optimizers. In light of stabilizing the decrease in the loss function, one can decrease after a number of epochs the value of the velocity parameter $k$ down to 0.1 in order to prevent high values in the objective function. \\
Now, we turn our attention to the results on the MNIST-Fashion dataset (the results are in the right part of the Table \ref{tab:MNIST_Results}). We can observe that \ref{NaG} with inertial parameter 0.5 achieves almost 0.875 accuracy value, taking into account that the step-size was chosen as 0.1. Further, \color{red} SSA 1 \color{black} and \color{red} SSA 2 \color{black} both achieve circa 0.92 accuracy on this test set. For lower values of the learning rate \ref{NaG} is better, but it seems that the hyper-parameters $k$ and $q$ compensate well into the velocity updates in order to achieve almost the same values of accuracy. The advantage of \ref{MiniBatch-SGD} and \ref{NaG} is that on the 20 epochs they achieve a lower value in the loss function, but after 50 the difference is negligible. Quite interestingly, in the case of the non-adaptive optimizers, Algorithm \ref{alg:SSA1-Ada} does not degrade after 20 epochs, even though MNIST-Fashion is harder to classify in comparison with the classical MNIST dataset. On the other hand, our adaptive algorithms seem to provide lower accuracy values at the end of the $100^{th}$ epoch and, in order to compensate, one can tune parameter $k$ to a greater value that will boost the value of the iterations. Finally, for RMSProp and Adam, we infer that they are not really suited for a step-size greater than $0.01$. Their advantage is that they do not overfit, which can be noticed from Table \ref{tab:MNIST_Results}, where the difference between the accuracy on training and test datasets appears imperceptible. \\

\begin{figure}[!ht]
     \centering
     \begin{subfigure}[b]{0.495\textwidth}
         \centering
         \includegraphics[width=\textwidth]{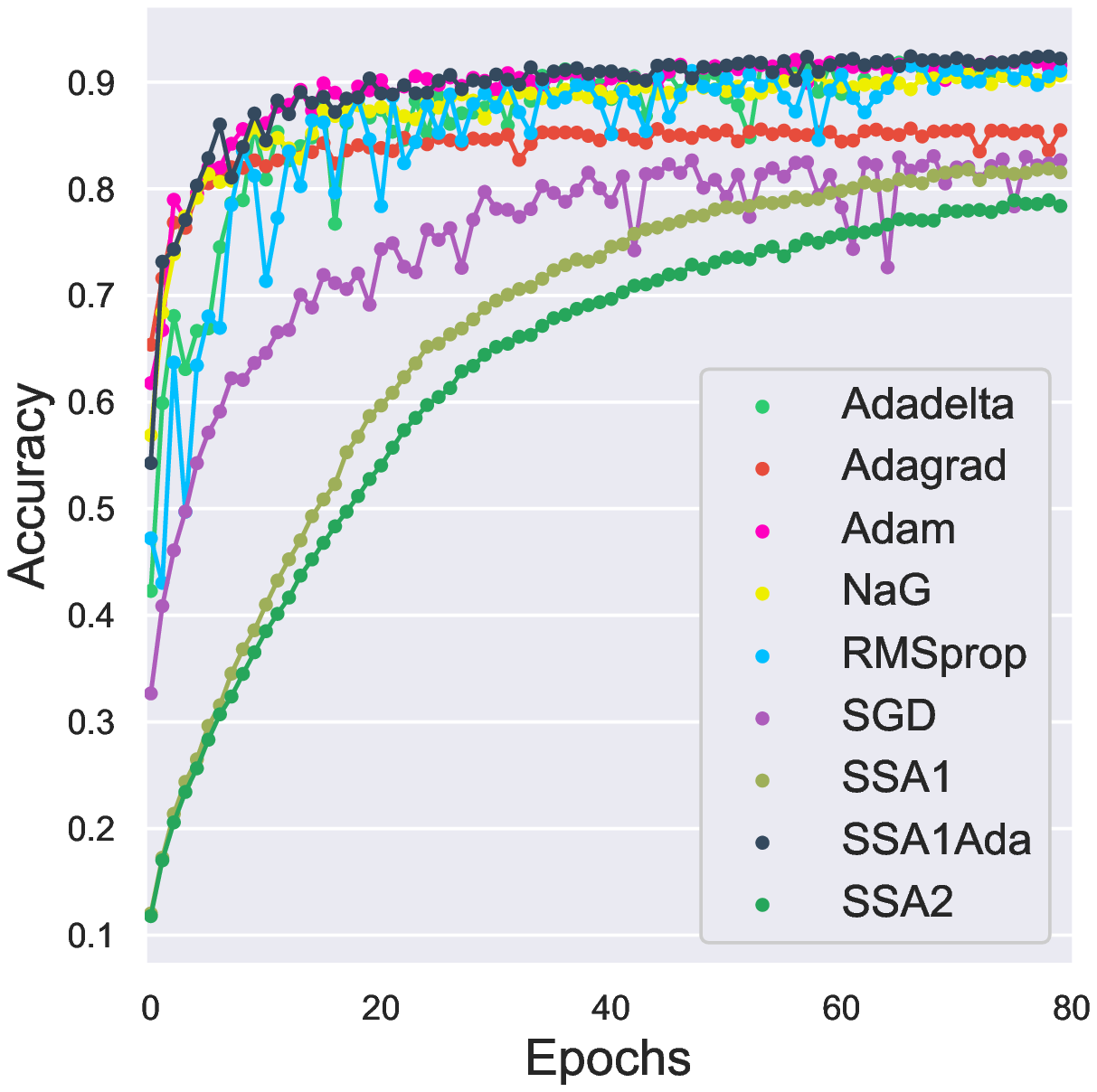}
         \caption{Accuracy comparison. Number of epochs 80.}
         \label{fig:CIFAR_60}
     \end{subfigure}
     \hfill
     \begin{subfigure}[b]{0.495\textwidth}
         \centering
         \includegraphics[width=\textwidth]{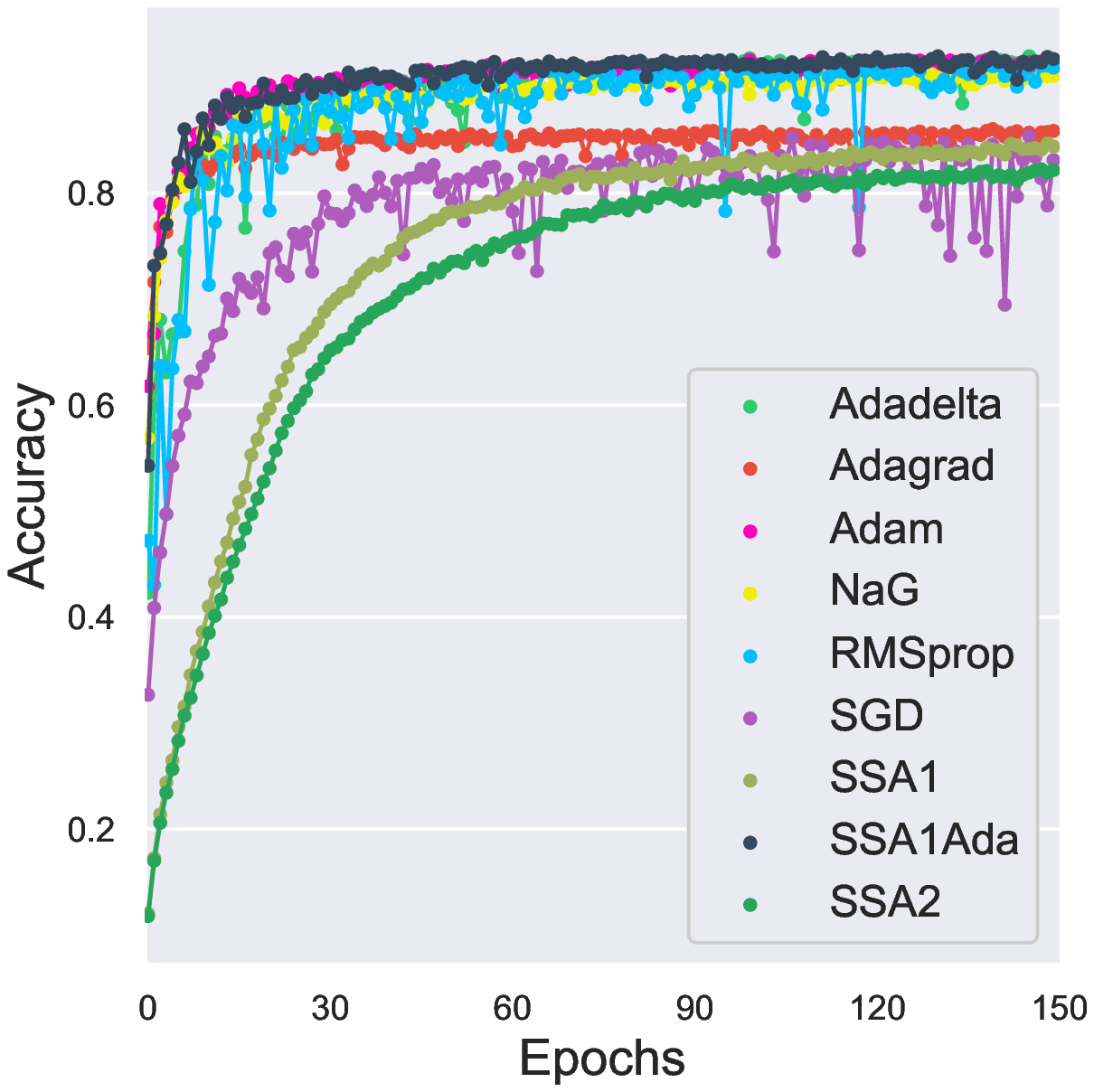}
         \caption{Accuracy comparison. Number of epochs 150.}
         \label{fig:CIFAR_150}
     \end{subfigure}
        \caption{Accuracy comparison for different optimizers on test CIFAR 10 dataset, using \textit{CrossEntropyLoss}.  Learning rates from table \ref{tab:CIFAR10_Results}.}
        \label{fig:CIFAR_acc_figures}
\end{figure}



Concerning Figure \ref{fig:CIFAR_60} and Figure \ref{fig:CIFAR_150}, we considered the comparison on CIFAR 10 test dataset for different optimization algorithms, using 80 epochs and 150 epochs, respectively. It can be derived that Algorithm \ref{alg:SSA1} does not present oscillations but they need more epochs in order to converge to a local minimum of the \textit{Cross Entropy Loss} function. Much better results are obtained by using adaptive variants, due to the tuning of the learning rate at each iteration on the mini-batches. It is worth noticing that the other adaptive algorithms RMSProp and Adam present natural oscillations determined by their aggressive step-size tuning. On the other hand, we recall that these results depend heavily on the randomization of weights and biases, since the local minimum of the loss function at which they converge depends on each and every optimizer. The full results are given in Table \ref{tab:CIFAR10_Results}. We ran our algorithms to 150 epochs and we have chosen the optimal step-size for each algorithm. \color{red} SSA 1 \color{black} and \color{red} SSA 2 \color{black} are both comparable to \ref{NaG} (with momentum 0.9), but in the first 20 epochs they present lower values in accuracy. On the other hand, RMSProp is comparable with Adadelta and Adagrad, but the latter one reaches lower values of accuracy after 20 epochs, due to the aggressive nature of the learning-rate tuning. Last but not least, we mention that in the last two lines of Table \ref{tab:CIFAR10_Results},  \color{red} SSA 1-Ada \color{black} yield better results compared to all the algorithms we have presented so far. Needless to say, a rigorous analysis of these optimization schemes entails numerous simulations, with different random values for the weights and biases of the neural network, in order to reach an average value for the accuracy on the test set. As far our splitting-based schemes go, the hyper-parameter $k$ play a major part, as in they must tuned by way of the validation set. This aims at achieving the lowest possible values in the chosen loss function.

\begin{remark}{\ \\}

It is interesting to notice that in Table \ref{tab:MNIST_Results} and in Table \ref{tab:CIFAR10_Results} we have the values for the accuracy on the train and test datasets for different optimizers. For our simulations we have used a fixed random seed. For practical purposes, one must generate different simulations on these datasets, since the convergence of these algorithms depend heavily on the initial random values of the weights and biases in the neural network.

\end{remark}

\begin{table}[!ht]
\centering
        \captionof{table}{Scores of the accuracy metric versus step-size and epochs for different optimizers on MNIST dataset (left) and on MNIST-Fashion dataset (right), using $NLLLoss$.}
  \label{tab:MNIST_Results}

\begin{tabular}{|| c | c | c | c || c | c | c | c ||}
\hline\hline
& \multicolumn{3}{c||}{MNIST} & \multicolumn{3}{c|}{MNIST-Fashion} & \\ \hline
\diagbox[width=5.5em]{$Optim.$}{$Epochs$} \cellcolor[gray]{0.9} & 20 \cellcolor[gray]{0.6} & 50 \cellcolor[gray]{0.6} & 100 \cellcolor[gray]{0.6} & 20 \cellcolor[gray]{0.6} & 50 \cellcolor[gray]{0.6} & 100 \cellcolor[gray]{0.6}& \emph{$h$} \cellcolor[gray]{0.9} \\ \hline

\multirow{3}{*}{SGD} & \diagbox[width=5em]{$0.9719$}{$0.9752$} & \diagbox[width=5em]{$0.9860$}{$0.9857$} & \diagbox[width=5em]{$0.9924$}{$0.9886$} & \diagbox[width=5em]{$0.8392$}{$0.8352$} & \diagbox[width=5em]{$0.8801$}{$0.8667$} & \diagbox[width=5em]{$0.9069$}{$0.8821$} & $10^{-3}$ \\ \cline{2-7}
& \diagbox[width=5em]{$0.9956$}{$0.9914$} & \diagbox[width=5em]{$0.9997$}{$0.9916$} & \diagbox[width=5em]{$1$}{$0.9914$} & \diagbox[width=5em]{$0.9187$}{$0.8996$} & \diagbox[width=5em]{$0.9666$}{$0.9111$} & \diagbox[width=5em]{$0.9972$}{$0.9142$}& $10^{-2}$ \\ \cline{2-7}
& \diagbox[width=5em]{$0.9999$}{$0.9934$} & \diagbox[width=5em]{$1$}{$0.994$} & \diagbox[width=5em]{$1$}{$0.9941$} & \diagbox[width=5em]{$0.9788$}{$0.9019$} & \diagbox[width=5em]{$1$}{$0.9175$} & \diagbox[width=5em]{$1$}{$0.9181$}& $10^{-1}$ \\\cline{2-7}

\hline \hline
\multirow{3}{*}{NaG (0.5)} & \diagbox[width=5em]{$0.9825$}{$0.9836$} & \diagbox[width=5em]{$0.9920$}{$0.9894$} & \diagbox[width=5em]{$0.9973$}{$0.9905$} &  \diagbox[width=5em]{$0.8687$}{$0.8590$} & \diagbox[width=5em]{$0.9036$}{$0.8889$} & \diagbox[width=5em]{$0.9330$}{$0.8987$} & $10^{-3}$\\ \cline{2-7}
& \diagbox[width=5em]{$0.9984$}{$0.9912$} & \diagbox[width=5em]{$1$}{$0.9919$} & \diagbox[width=5em]{$1$}{$0.992$} & \diagbox[width=5em]{$0.9419$}{$0.9124$} & \diagbox[width=5em]{$0.9913$}{$0.9162$} & \diagbox[width=5em]{$1$}{$0.9176$} & $10^{-2}$\\ \cline{2-7} 
& \diagbox[width=5em]{$1$}{$0.9944$} & \diagbox[width=5em]{$1$}{$0.9942$} & \diagbox[width=5em]{$1$}{$0.9942$} &  \diagbox[width=5em]{$0.9715$}{$0.9003$} & \diagbox[width=5em]{$0.9836$}{$0.8874$} & \diagbox[width=5em]{$0.9687$}{$0.8800$} & $10^{-1}$\\  \hline

\hline \hline

\multirow{3}{*}{SSA 1} & \diagbox[width=5em]{$0.327$}{$0.9396$} & \diagbox[width=5em]{$0.9783$}{$0.9806$} & \diagbox[width=5em]{$0.9891$}{$0.9881$} & \diagbox[width=5em]{$0.7780$}{$0.7666$} & \diagbox[width=5em]{$0.8574$}{$0.8488$} & \diagbox[width=5em]{$0.8939$}{$0.8776$} & $10^{-3}$ \\ \cline{2-7}
& \diagbox[width=5em]{$0.9938$}{$0.9907$} & \diagbox[width=5em]{$0.9996$}{$0.9911$} & \diagbox[width=5em]{$1$}{$0.9909$} & \diagbox[width=5em]{$0.9131$}{$0.8973$} & \diagbox[width=5em]{$0.9619$}{$0.9119$} & \diagbox[width=5em]{$0.9976$}{$0.9108$} & $10^{-2}$\\ \cline{2-7} 
& \diagbox[width=5em]{$1$}{$0.9930$} & \diagbox[width=5em]{$1$}{$0.9935$} & \diagbox[width=5em]{$1$}{$0.9934$} &  
\diagbox[width=5em]{$0.9802$}{$0.9082$} & \diagbox[width=5em]{$0.9972$}{$0.9156$} & \diagbox[width=5em]{$1$}{$0.9161$} & $10^{-1}$\\ \hline

\hline \hline

\multirow{3}{*}{SSA 2} & \diagbox[width=5em]{$0.9173$}{$0.9218$} & \diagbox[width=5em]{$0.9688$}{$0.9734$} & \diagbox[width=5em]{$0.9838$}{$0.9846$} &  \diagbox[width=5em]{$0.7604$}{$0.7531$} & \diagbox[width=5em]{$0.8366$}{$0.8264$} & \diagbox[width=5em]{$0.8735$}{$0.8608$} & $10^{-3}$ \\ \cline{2-7}
& \diagbox[width=5em]{$0.9897$}{$0.9873$} & \diagbox[width=5em]{$0.9975$}{$0.9910$} & \diagbox[width=5em]{$0.9998$}{$0.9910$} & \diagbox[width=5em]{$0.8953$}{$0.8811$} & \diagbox[width=5em]{$0.9344$}{$0.9087$} & \diagbox[width=5em]{$0.9740$}{$0.9112$} & $10^{-2}$\\ \cline{2-7} 
& \diagbox[width=5em]{$0.9997$}{$0.9915$} & \diagbox[width=5em]{$1$}{$0.9924$} & \diagbox[width=5em]{$1$}{$0.9923$} & \diagbox[width=5em]{$0.9678$}{$0.9116$} & \diagbox[width=5em]{$1$}{$0.9184$} & \diagbox[width=5em]{$1$}{$0.9175$} & $10^{-1}$ \\  \hline \hline

\multirow{3}{*}{RMSProp} & \diagbox[width=5em]{$0.9989$}{$0.9904$} & \diagbox[width=5em]{$0.9994$}{$0.9908$} & \diagbox[width=5em]{$0.9998$}{$0.9921$} & \diagbox[width=5em]{$0.9848$}{$0.9096$} & \diagbox[width=5em]{$0.9933$}{$0.9042$} & \diagbox[width=5em]{$0.9960$}{$0.9001$} & $10^{-3}$\\ \cline{2-7}
& \diagbox[width=5em]{$0.1107$}{$0.1135$} & \diagbox[width=5em]{$0.1111$}{$0.1028$} & \diagbox[width=5em]{$0.1098$}{$0.1135$} & \diagbox[width=5em]{$0.1002$}{$0.1$} & \diagbox[width=5em]{$0.0999$}{$0.1$} & \diagbox[width=5em]{$0.0991$}{$0.1$} & $10^{-2}$\\ \cline{2-7} 
& \diagbox[width=5em]{$0.1009$}{$0.1135$} & \diagbox[width=5em]{$0.1031$}{$0.1028$} & \diagbox[width=5em]{$0.1040$}{$0.1028$} & \diagbox[width=5em]{$0.0999$}{$0.1$} & \diagbox[width=5em]{$0.1018$}{$0.1$} & \diagbox[width=5em]{$0.0979$}{$0.1$} & $10^{-1}$\\  \hline \hline

\multirow{3}{*}{Adam} & \diagbox[width=5em]{$0.9985$}{$0.9921$} & \diagbox[width=5em]{$0.9988$}{$0.9909$} & \diagbox[width=5em]{$0.9996$}{$0.9907$} & \diagbox[width=5em]{$0.9853$}{$0.9108$} & \diagbox[width=5em]{$0.9933$}{$0.9063$} & \diagbox[width=5em]{$0.9966$}{$0.9081$} & $10^{-3}$ \\ \cline{2-7}
& \diagbox[width=5em]{$0.1108$}{$0.1135$} & \diagbox[width=5em]{$0.1105$}{$0.1028$} & \diagbox[width=5em]{$0.1100$}{$0.1135$} & \diagbox[width=5em]{$0.8692$}{$0.8517$} & \diagbox[width=5em]{$0.8726$}{$0.8544$} & \diagbox[width=5em]{$0.8632$}{$0.8335$} & $10^{-2}$\\ \cline{2-7} 
& \diagbox[width=5em]{$0.1031$}{$0.1032$} & \diagbox[width=5em]{$0.1028$}{$0.1028$} & \diagbox[width=5em]{$0.1044$}{$0.1135$} &  \diagbox[width=5em]{$0.1$}{$0.1$} & \diagbox[width=5em]{$0.1005$}{$0.1$} & \diagbox[width=5em]{$0.0998$}{$0.1$} & $10^{-1}$\\  

\hline \hline

\multirow{1}{*}{Adadelta} & \diagbox[width=5em]{$1$}{$0.9945$} & \diagbox[width=5em]{$1$}{$0.9945$} & \diagbox[width=5em]{$1$}{$0.9944$} & \diagbox[width=5em]{$0.9847$}{$0.9054$} & \diagbox[width=5em]{$0.9978$}{$0.9064$} & \diagbox[width=5em]{$1$}{$0.9079$} & $1$ \\ \hline \hline

\multirow{1}{*}{SSA 1 - Ada} & \diagbox[width=5em]{$0.9969$}{$0.9897$} & \diagbox[width=5em]{$0.9965$}{$0.9883$} & \diagbox[width=5em]{$0.0987$}{$0.0980$} & \diagbox[width=5em]{$0.9383$}{$0.8840$} & \diagbox[width=5em]{$0.9620$}{$0.8625$} & \diagbox[width=5em]{$0.1$}{$0.1$} & $1$  \\ \hline \hline

\end{tabular}

\end{table}

\clearpage

\begin{table}[!ht]
\centering
        \captionof{table}{Scores of the accuracy metric versus step-size and epochs for different optimizers on CIFAR 10 dataset, using \emph{Cross Entropy Loss}.}
    \label{tab:CIFAR10_Results}

\begin{tabular}{|| c | c | c | c | c | c ||}
\hline\hline
\diagbox[width=5.5em]{$Optim.$}{$Epochs$} \cellcolor[gray]{0.9} & 20 \cellcolor[gray]{0.6} & 50 \cellcolor[gray]{0.6} & 100 \cellcolor[gray]{0.6} & 150 \cellcolor[gray]{0.6} & \emph{$h$} \cellcolor[gray]{0.9} \\ \hline

\multirow{1}{*}{SGD} & \diagbox[width=5em]{$0.7747$}{$0.6913$} & \diagbox[width=5em]{$0.9168$}{$0.8084$} & \diagbox[width=5em]{$0.9792$}{$0.8350$} & \diagbox[width=5em]{$0.9907$}{$0.8313$} & $10^{-3}$  \\  

\hline \hline

\multirow{1}{*}{NaG (0.9)} & \diagbox[width=5em]{$0.9459$}{$0.8727$} & \diagbox[width=5em]{$0.9895$}{$0.8934$} & \diagbox[width=5em]{$0.9978$}{$0.8932$} & \diagbox[width=5em]{$0.9991$}{$0.9108$} & $10^{-3}$ \\ 

\hline \hline

\multirow{1}{*}{SSA 1} & \diagbox[width=5em]{$0.5736$}{$0.5868$} & \diagbox[width=5em]{$0.8248$}{$0.7806$} & \diagbox[width=5em]{$0.9521$}{$0.8302$} & \diagbox[width=5em]{$0.9833$}{$0.8440$}& $10^{-3}$ \\

\hline \hline

\multirow{1}{*}{SSA 2} & \diagbox[width=5em]{$0.5151$}{$0.5278$} & \diagbox[width=5em]{$0.7570$}{$0.7312$} & \diagbox[width=5em]{$0.9042$}{$0.8071$} & \diagbox[width=5em]{$0.9605$}{$0.8218$} & $10^{-3}$ \\  \hline \hline

\multirow{1}{*}{RMSProp} & \diagbox[width=5em]{$0.9463$}{$0.8464$} & \diagbox[width=5em]{$0.9844$}{$0.8930$} & \diagbox[width=5em]{$0.9936$}{$0.9173$} & \diagbox[width=5em]{$0.9956$}{$0.9185$} & $10^{-3}$ \\  \hline \hline

\multirow{1}{*}{Adam} & \diagbox[width=5em]{$0.9557$}{$0.8924$} & \diagbox[width=5em]{$0.9874$}{$0.9150$} & \diagbox[width=5em]{$0.9940$}{$0.9256$} & \diagbox[width=5em]{$0.9966$}{$0.9254$} & $10^{-3}$ \\

\hline \hline

\multirow{1}{*}{Adagrad} & \diagbox[width=5em]{$0.9325$}{$0.8388$} & \diagbox[width=5em]{$0.9841$}{$0.8508$} & \diagbox[width=5em]{$0.9949$}{$0.8507$} & \diagbox[width=5em]{$0.9973$}{$0.8588$} & $10^{-3}$ \\ \hline \hline

\multirow{1}{*}{Adadelta} & \diagbox[width=5em]{$0.9531$}{$0.8669$} & \diagbox[width=5em]{$0.9903$}{$0.9084$} & \diagbox[width=5em]{$0.9975$}{$0.9273$} & \diagbox[width=5em]{$0.9986$}{$0.9258$} & $1$ \\ \hline \hline

\multirow{1}{*}{SSA 1 - Ada} & \diagbox[width=5em]{$0.9540$}{$0.9035$} & \diagbox[width=5em]{$0.9904$}{$0.9118$} & \diagbox[width=5em]{$0.9967$}{$0.9219$} & \diagbox[width=5em]{$0.9979$}{$0.9267$} & $1$ \\ \hline \hline

\end{tabular}

\end{table}

\begin{remark}{\ \\}

In the \ref{NaG} implementation from \texttt{Pytorch}, the momentum coefficient $\beta_n$ is set to a fixed value between $0.5$ and $0.9$. Thus, for a smooth implementation of our optimization schemes, one can also use a fixed inertial coefficient, instead of the non-constant sequence, defined as $\beta_n = \tfrac{n}{n+3}$ or $\beta_n = \tfrac{n-1}{n+2}$ (see Remark \ref{R1}),  where $n$ represents the iteration employed in the computation of the gradient on the current mini-batch.

\end{remark}

\begin{remark}{\ \\}

The algorithms \color{red} SSA 1 \color{black}, \color{red} SSA 2 \color{black} and \color{red} SSA 1-Ada \color{black} are based upon the hyper-parameter $k$, with the default value of 2.0. The increase of the loss function values and the decrease of the accuracy on the test dataset, call for the tuning $k$ with respect to the current iteration, where the gradient of the mini-batches is computed. For example, if the accuracy does not change from epoch to epoch, then one can boost the value of $k$ from 2.0 to 10.0. On the flip side, if the chances of overfitting are high, then one can tune $k$ from 2.0 down to 0.1. The parameter $k$ can be non-constant, i.e. $k = k(n)$. If the algorithm indeed overfits, one can set it to decay exponentially, such as $k(n) = e^{k (-1/n)}$, where $k$ stands for the default value.
\end{remark}

At last, we present a crucial remark concerning the adaptive counterpart of the first splitting algorithm.

\begin{remark}{\ \\}\label{LastRemark}
A method for improving \color{red} SSA 1-Ada \color{black}, given by Algorithm \ref{alg:SSA1-Ada} is to use the current value of the gradient $\nabla f_{\theta}(z_{n+1} ; x^{(i:i+m)}, y^{(i:i+m)})$ and the squared gradient $\nabla^2 f_{\theta}(z_{n+1} ; x^{(i:i+m)}, y^{(i:i+m)})$ at the intermediate point $z_{n+1} = \theta_n + h \beta_n v_n$. A simple way for implementing this is to compute the additional iteration $z_{n+1}$ before the accumulation of gradient and before the computation updates. Simply put, this means moving line 18 after line 9 in the algorithm and then use the gradient and the squared gradient in this newly updated iteration.
\end{remark}

Now, before our next remark consisting of some explanations behind the training time of the neural network, we emphasize the fact that \color{red} SSA1-Ada \color{black} can be implemented as in Algorithm \ref{alg:SSA1-Ada} or as specified in Remark \ref{LastRemark}. The main idea is that different constructions of the adaptive splitting-based algorithms represent only a guideline for deep neural network implementations of \color{red} SSA1-Ada \color{black}.

\begin{remark}{\ \\}
Now, we present some facts about the training time regarding the machine learning algorithms. As above, we have considered the three datasets, namely MNIST, MNIST-Fashion and CIFAR-10. On each of them we have compared the time to train the neural network with the same optimizers as before: Adadelta, Adam, RMSProp, \ref{NaG}, SGD, \color{red} SSA1 \color{black}, \color{red} SSA2 \color{black} and \color{red} SSA 1-Ada \color{black}. The results are presented in Table \ref{tab:CMNIST_Time}, Table \ref{tab:FASH_Time} and Table \ref{tab:CIFAR10_Time}, respectively. The procedure is in a sense statistical: we have stored the training time on 100 epochs and then we have constructed an array consisting of the 100 simulations. As an implementation we have used the classical \texttt{Pandas} toolbox from Python, in which we have stored the mean, the standard deviation, the minimum value, the maximum value, the quantiles and the sum of all these values, for the array containing the results of the training time corresponding to 100 epochs. Now, we will briefly present the comparative results for each of the datasets. In Table \ref{tab:CMNIST_Time} we observe that \color{red} SSA1 \color{black} and \color{red} SSA2 \color{black} present an mean value close to the other algorithms. The standard deviation is, on average, as the same as Adadelta and RMSProp. In this case only SGD has a much lower value in the standar deviation, suggesting that is more stable than the other optimizers. The maximum values suggests also that our algorithms stand close to Adadelta, Adam and \ref{NaG}. On the other hand, we point out that adaptive \color{red} SSA1-Ada \color{black} has values that are close enough to Adadelta. This is linked to the fact that both Adadelta and our algorithm are adaptive counterparts of inertial optimizers. Now, we turn our attention to Table \ref{tab:FASH_Time}. In the case of the MNIST-Fashion dataset, the mean of the algorithms is lower in comparison with the case of the classical MNIST. Furthermore, the behavior of \color{red} SSA1 \color{black} and \color{red} SSA2 \color{black} is similar to the previous experiment in the case of the mean and of the standard deviation. This is connected to the fact that the same network architecture was used in the case of both datasets. Finally, we will present the computational results regarding Table \ref{tab:CIFAR10_Time}. In the case of complex architectures like that used in CIFAR-10 dataset, our splitting algorithms have a high standard deviation, which is in resonance with the instability regarding the training time over a long period of epochs. Moreover, the mean of \color{red} SSA1 \color{black} and \color{red} SSA2 \color{black} are close to the values of Adadelta and Adagrad. Also, it seems that \ref{NaG} and SGD are faster than our splitting algorithms. Now, the adaptive \color{red} SSA1-Ada \color{black} requires much more training time and has a higher variance in comparison with the other stochastic algorithms. Even though this algorithm requires a greater training time, the advantage is that it compensates in accuracy and it does not lead to overfitting (see the previous tables). Finally, we point out that, in order to make \color{red} SSA1-Ada \color{black} much faster, one can employ the Remark \ref{LastRemark}, where we have discussed a different implementation. This will lead to a faster algorithm that encompasses both the inertial effects and also the stability in the loss function.
\end{remark}
\vspace{-0.2cm}
\begin{table}[!ht]
\centering
        \captionof{table}{MNIST execution time.}
    \label{tab:CMNIST_Time}

\begin{tabular}{|| c | c | c | c | c | c | c | c | c ||}
\hline\hline
\diagbox[width=4.5em]{$Alg$}{$Stat$} \cellcolor[gray]{0.9} & mean & std & min & 25\% & 50\% & 75\%& max & sum \\ \hline

\multirow{1}{*}{Adadelta \cellcolor[gray]{0.6}} & $10.25$ & $1.51$ & $7.63$ & $8.97$ & $10.45$ & $11.44$ &$12.96$ & $1025.33$\\  

\hline \hline

\multirow{1}{*}{Adam \cellcolor[gray]{0.6}} & $10.22$ & $1.70$ & $7.39$ & $8.80$ &$10.06$ & $11.55$ &$14.48$ & $1022.28$\\  

\hline \hline

\multirow{1}{*}{RMSProp \cellcolor[gray]{0.6}} & $9.14$ & $1.58$ & $7.55$ & $7.73$ & $8.51$ & $10.33$ & 
$12.82$ & $914.26$\\  

\hline \hline

\multirow{1}{*}{NaG \cellcolor[gray]{0.6}} & $10.24$ & $1.82$ & $7.44$ & $8.81$ & $10.08$ & $11.80$ &
$14.40$ & $1024.89$\\  

\hline \hline

\multirow{1}{*}{SGD \cellcolor[gray]{0.6}} & $8.39$ & $1.12$ & $7.52$ & $7.73$ &$7.89$ &
$8.69$ &$12.63$ & $839.54$\\  
						
\hline \hline

\multirow{1}{*}{SSA1 \cellcolor[gray]{0.6} } & $10.01$ & $1.53$ & $7.61$ & $8.58$ &$10.11$ &
$11.25$ &$13.29$ & $1001.22$\\  

\hline \hline

\multirow{1}{*}{SSA2 \cellcolor[gray]{0.6}} & $9.57$ & $1.50$ & $7.48$ & $8.20$ &$9.39$ &
$10.64$ &$13.99$ & $957.61$\\  

\hline \hline

\multirow{1}{*}{SSA1-Ada \cellcolor[gray]{0.6}} & $11.61$ & $1.42$ & $8.83$ & $10.62$ &$11.54$ & 
$12.58$ &$15.40$ & $1161.33$\\

\hline \hline

\end{tabular}

\end{table}

\begin{table}
\centering
        \captionof{table}{MNIST-Fashion execution time.}
    \label{tab:FASH_Time}

\begin{tabular}{|| c | c | c | c | c | c | c | c | c ||}
\hline\hline
\diagbox[width=4.5em]{$Alg$}{$Stat$}  \cellcolor[gray]{0.9} & mean  & std  & min & 25\%  & 50\%  & 75\%  & max & sum  \\ \hline

\multirow{1}{*}{Adadelta} \cellcolor[gray]{0.6}& $9.84$ & $1.62$ & $7.59$ & $8.29$ & $9.85$ &
$11.03$ &$13.96$ & $984.44$\\  

\hline \hline

\multirow{1}{*}{Adam}\cellcolor[gray]{0.6} & $10.25$ & $1.77$ & $7.48$ & $8.62$ & $10.14$ & $11.67$ &$14.18$ & $1025.30$\\  

\hline \hline

\multirow{1}{*}{RMSProp}\cellcolor[gray]{0.6} & $10.03$ & $1.93$ & $7.49$ & $8.29$ & $9.53$ & $11.44$ &$14.49$ & $1003.04$\\  

\hline \hline

\multirow{1}{*}{NaG} \cellcolor[gray]{0.6}& $9.83$ & $1.73$ & $7.60$ & $8.40$ & $9.58$ & $10.84$ &$14.26$ & $983.52$\\  

\hline \hline

\multirow{1}{*}{SGD} \cellcolor[gray]{0.6}& $8.15$ & $0.87$ & $7.46$ & $7.65$ & $7.82$ &
$8.43$ &$13.37$ & $815.31$\\  

\hline \hline

\multirow{1}{*}{SSA1} \cellcolor[gray]{0.6}& $9.68$ & $1.52$ & $7.65$ & $8.58$ & $9.23$ & $10.47$ &$13.75$ & $968.59$\\  

\hline \hline

\multirow{1}{*}{SSA2}\cellcolor[gray]{0.6} & $9.62$ & $1.61$ & $7.47$ & $8.12$ & $9.65$ &
$10.69$ &$13.22$ & $962.08$\\  

\hline \hline

\multirow{1}{*}{SSA1-Ada} \cellcolor[gray]{0.6}& $11.70$ & $1.25$ & $8.85$ & $10.93$ & $11.77$ &
$12.62$ &$14.52$ & $1170.42$\\

\hline \hline

\end{tabular}
\end{table}


\begin{table}[!ht]
\centering
        \captionof{table}{CIFAR execution time.}
    \label{tab:CIFAR10_Time}

\begin{tabular}{|| c | c | c | c | c | c | c | c | c |||}
\hline\hline
\diagbox[width=4.5em]{$Alg$}{$Stat$} \cellcolor[gray]{0.9} & mean & std & min & 25\% & 50\%& 75\% & max & sum \\ \hline

\multirow{1}{*}{Adadelta \cellcolor[gray]{0.6} } & $104.74$ & $0.99$ & $102.65$ & $104.27$ & $104.60$ & $105.18$ & 
$110.52$ &$10474.36$ \\  
\hline \hline

\multirow{1}{*}{Adagrad \cellcolor[gray]{0.6}} & $91.82$ & $0.42$ & $91.39$ & $91.47$ & $91.72$ & $92.07$ & 
$93.11$ &$9182.64$ \\
\hline \hline

\multirow{1}{*}{Adam \cellcolor[gray]{0.6}} & $97.30$ & $0.58$ & $96.14$ & $96.87$ & $97.17$ & $97.56$ & 
$98.80$ &$9730.21$ \\  
\hline \hline

\multirow{1}{*}{RMSProp \cellcolor[gray]{0.6}} & $94.52$ & $0.50$ & $93.11$ & $94.17$ & $94.40$ & $94.83$ & 
$96.21$ &$9452.95$ \\  
\hline \hline

\multirow{1}{*}{NaG \cellcolor[gray]{0.6} } & $91.00$ & $0.46$ & $90.43$ & $90.61$ & $90.92$ & $91.16$ & 
$92.73$ &$9100.56$ \\  
\hline \hline

\multirow{1}{*}{SGD \cellcolor[gray]{0.6}} & $85.92$ & $0.29$ & $85.55$ & $85.68$ & $85.84$ & $86.12$ & 
$87.18$ &$8592.01$ \\  

\hline \hline

\multirow{1}{*}{SSA1 \cellcolor[gray]{0.6}} & $112.23$ & $2.25$ & $107.88$ & $110.18$ & $112.33$ & $114.33$ & 
$115.79$ &$11223.46$ \\  

\hline \hline

\multirow{1}{*}{SSA2 \cellcolor[gray]{0.6}} & $106.81$ & $2.55$ & $102.76$ & $104.63$ & $105.65$ & $109.61$ & 
$110.96$ &$10681.89$ \\  
\hline \hline
\multirow{1}{*}{SSA1-Ada \cellcolor[gray]{0.6}} & $179.89$ & $4.54$ & $168.86$ & $176.53$ & $179.98$ & $183.60$ & 
$186.96$ &$17989.26$ \\  

\hline \hline

\end{tabular}

\end{table}

\begin{figure}[!ht]
     \centering
     \begin{subfigure}[b]{0.49\textwidth}
         \centering
         \includegraphics[width=\textwidth]{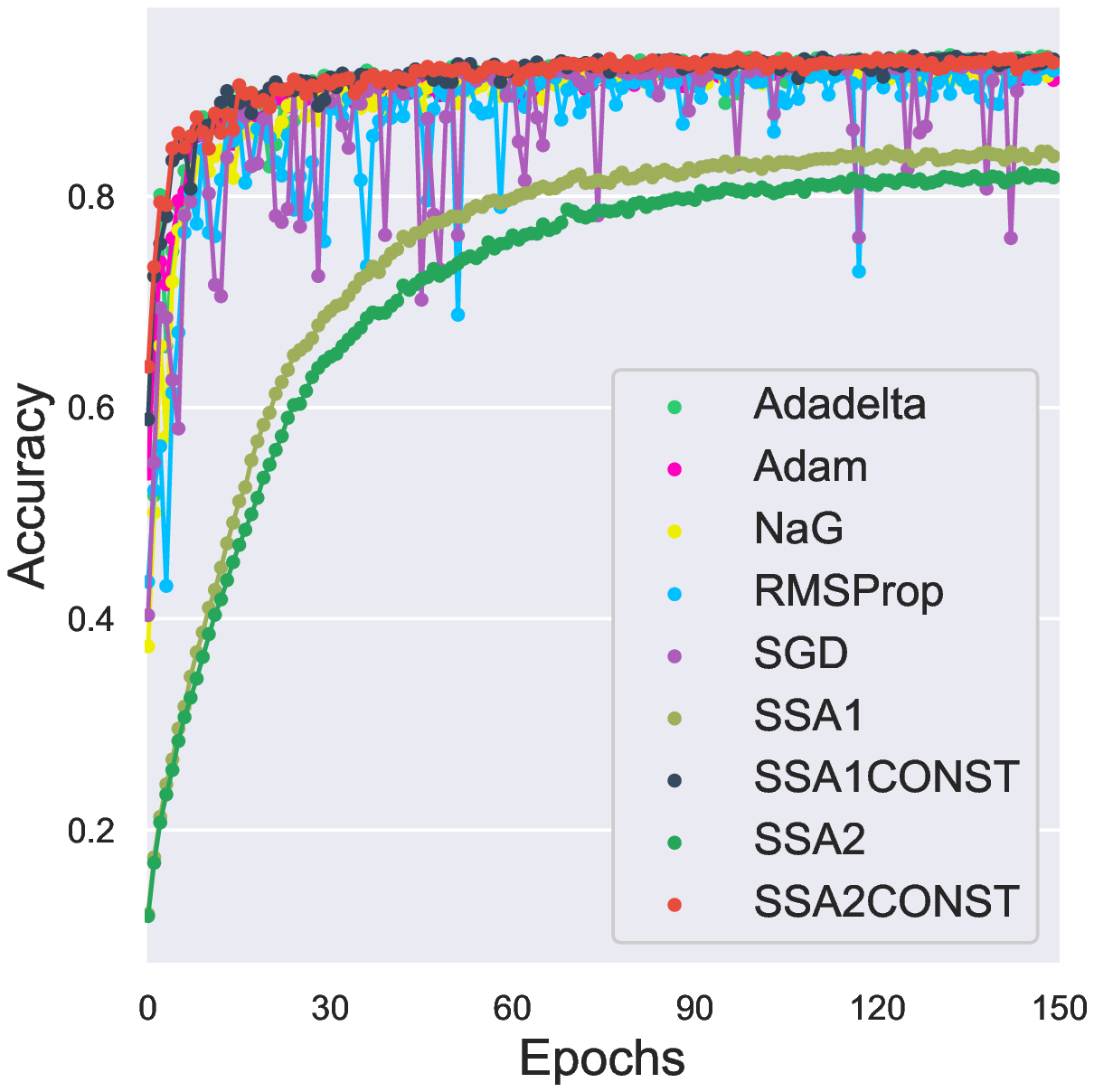}
         \caption{Accuracy comparison.}
         \label{fig:ComparisonCIFAR_CONST_Acc}
     \end{subfigure}
     \hfill
     \begin{subfigure}[b]{0.49\textwidth}
         \centering
         \includegraphics[width=\textwidth]{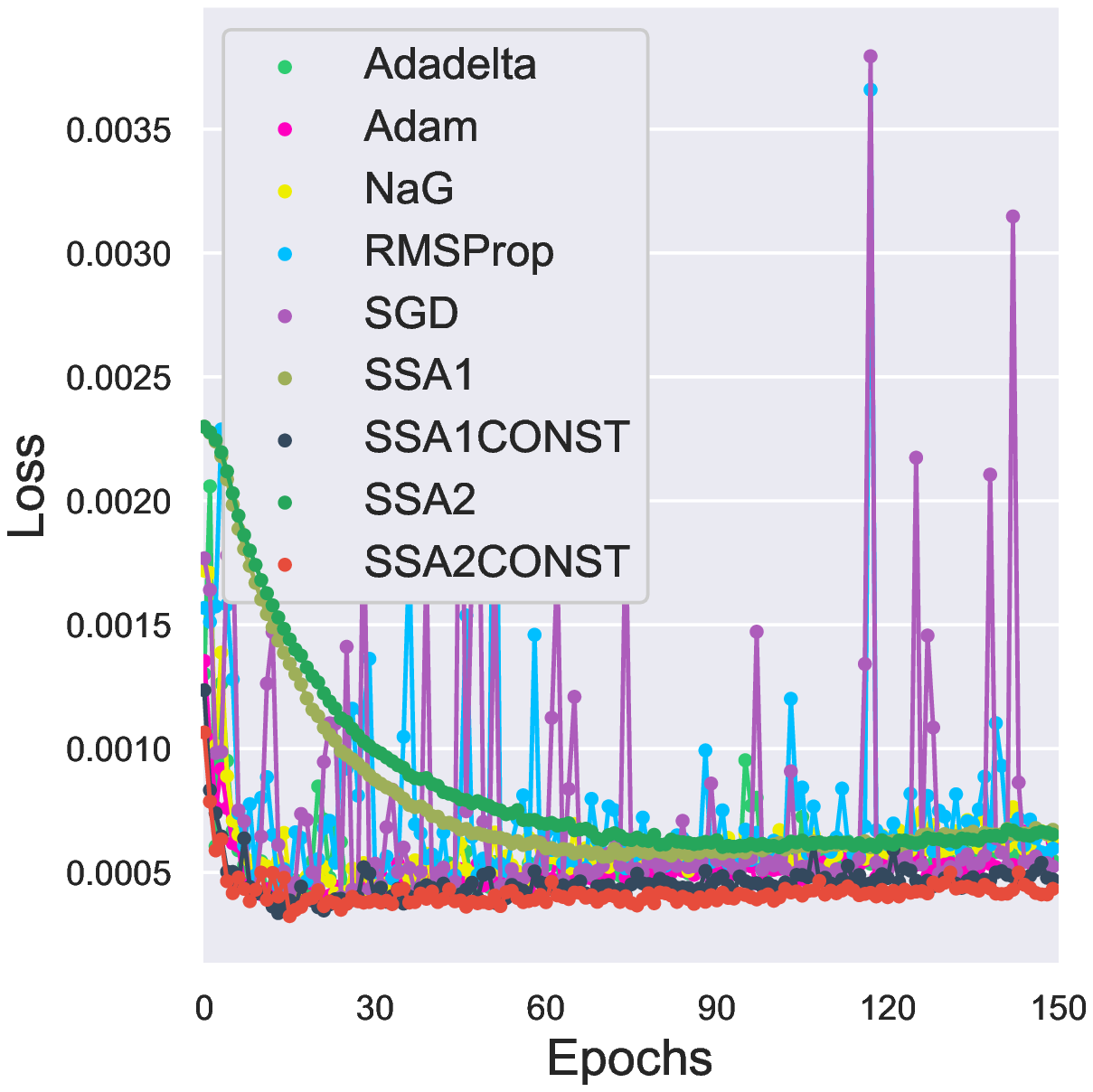}
         \caption{Comparison for the decrease in the loss function.}
         \label{fig:ComparisonCIFAR_CONST_Loss}
     \end{subfigure}
        \caption{Numerical experiments on the CIFAR 10 dataset for adaptive and non-adaptive algorithms, respectively. Number of epochs 150.}
        \label{fig:CIFAR_CONST}
\end{figure}

\begin{remark}\label{RemarkCIFARConst}

Nesterov algorithm \ref{NaG} developed in \cite{Nesterov} has been analyzed theoretically in different areas of the optimization community. The algorithm is based upon the inertial coefficient given as $\beta_n = \tfrac{n-1}{n+2}$ or $\beta_n = \tfrac{n}{n+3}$, as in Remark \ref{R1}. From a practical point of view, in \texttt{PyTorch} \footnote{ \url{https://pytorch.org/docs/stable/optim.html}} and in \texttt{Tensorflow} \footnote{ \url{https://www.tensorflow.org/api_docs/python/tf/compat/v1/train/MomentumOptimizer} }, the momentum coefficient $\beta_n$ is considered as a constant, with values between $0.5$ and $0.9$. The implementation of the momentum (see \cite{Sutskever}) as a constant term $\beta_n = \beta$ gives a boost in the empirical rate of convergence, due to the fact that $\lim\limits_{n \to \infty} \beta_n = 1$. Hence, we have the natural theoretical interpretation of \ref{NaG} with $\beta_n = \tfrac{n-1}{n+2}$ or $\beta_n = \tfrac{n}{n+3}$ and the practical version, when $\beta_n = \beta$. We gave a theoretical motivation for both our splitting-based algorithms with a nonconstant inertial sequence $(\beta_n)_{n \in \mathbb{N}}$ and we have compared with adaptive or non-adaptive algorithms (e.g. \ref{NaG} and \ref{MiniBatch-SGD}). Now, taking into account the above explanation, we have considered in Figure \ref{fig:ComparisonCIFAR_CONST_Acc} and in Figure \ref{fig:ComparisonCIFAR_CONST_Loss}, some numerical experiments on the CIFAR 10 test dataset, over 150 epochs. We have considered the following algorithms: Adadelta (with constant learning $1.0$), Adam and RMSprop algorithms (with learning rate $0.003$), SGD and Nesterov (with learning rate $0.5$) and both \color{red} SSA1 \color{black} and \color{red} SSA2 \color{black}, with $0.001$. Furthermore, the constant momentum term of Nesterov was set to $0.5$ as in the case of the constant variants of our algorithms (also for these algorithms we have set the learning rate to $0.5$). We observe, that Adam and RMSprop are unstable with respect to the learning rate due to their inherent adaptativity. On the other hand, we have chosen a larger learning rate for the constant momentum variants of both \color{red} SSA1 \color{black} and \color{red} SSA2 \color{black}, in order to show the potential smoothing effect available also for high values of the learning rate, with respect to the optimizers oscillations (due to the stochasticity). \\
We show that, similar to the computational implementation of \ref{NaG}, the constant momentum algorithms \color{red} SSA 1 \color{black} and \color{red} SSA2 \color{black} admit a better rate of convergence when the momentum coefficient is $\beta_n = \beta \in [0.5, 0.9]$. Furthermore, we can observe that our algorithms have a faster rate of convergence compared with all the algorithms we have considered so far (these algorithms have the name \color{red} \textit{SSA1 CONST} \color{black} and \color{red} \textit{SSA2 CONST} \color{black}, respectively), in the sense that we have obtained a better accuracy. The comparison reveals two computational aspects: \\
As in the case of \ref{NaG}, we have constructed a non-adaptive optimizer with a momentum term $\beta_n = \tfrac{n-1}{n+2}$ or $\beta_n = \tfrac{n}{n+3}$. From a theoretical point of view, our algorithms resemble \ref{MiniBatch-SGD} and \ref{NaG} where $\beta_n$ depends on the iteration values, but from a practical point of view, we have considered that a fair comparison needs to be made when $\beta_n = \beta$ with values between $0.5$ and $0.9$ for both \color{red} SSA1 \color{black} and \color{red} SSA2 \color{black}, and also for SGD and Nesterov accelerated gradient method (see also the implementation of momentum method in \texttt{PyTorch} as reminded above). In this case, obviously the underlying dynamical systems will simplify and will contain a constant damping term, but our purpose of the present remark is to give a fundamental approach that gives us a good and practical behavior related to speeding up the empirical convergence rate of the optimizers. \\
Last but not least, we mention that our algorithms have a similar property to non-adaptive optimizers like SGD and \ref{NaG}, in the sense that they generalize better than adaptive algorithms (see for example \cite{Wilson}). This can be seen in Figure \ref{fig:ComparisonCIFAR_CONST_Acc} and in Figure \ref{fig:ComparisonCIFAR_CONST_Loss}, respectively. We mention that although the property that we have mentioned above is present for both inertial algorithms and also for \color{red} SSA1 \color{black} and \color{red} SSA2 \color{black}, our algorithms are not inertial optimizers, but they are multi-step type methods since they contain both $y_n$ and $y_{n-1}$. For example, \ref{SeqSplitt-I-Alg} can be written as follows:
\begin{align*}
\begin{cases}
y_n = u_n + \beta_n \beta_{n-1}^k (u_n - u_{n-1}) + \beta_n \beta_{n-1}^{k-1}(1-h \beta_{n-1})(1-\beta_{n-1}^2)(y_{n-1}-u_{n-1}) \\
u_{n+1} = u_n + \beta_n (1-h \beta_n)(y_n - u_n) - h^2 \nabla f(y_n)
\end{cases}
\end{align*}
Similarly, \ref{SeqSplitt-II-Alg} can be reduced to the following alternate form:
\begin{align*}
\begin{cases}
y_n = u_n + \beta_n \beta_{n-1}^k (u_n - u_{n-1}) - h^2 \beta_n \beta_{n-1}^k \nabla f(y_{n-1}) \\ \\
u_{n+1} = u_n + \dfrac{1-h \beta_n}{\beta_n} (y_n - u_n)
\end{cases}    
\end{align*}
Finally, we emphasize the fact that one can reach the above algorithms (equivalent with \color{red} SSA1 \color{black} and \color{red} SSA2 \color{black}), using the value of discrete velocity, i.e. this consists in expressing $v_{n+1}$ in terms of $u_{n+1}$, $u_n$, $v_n$ and $y_n$.
\color{black}
\end{remark}

\section{Discussions}\label{Discussions_Section}

In this paper we have introduced new types of optimization algorithms that are competitive in the neural network training with the momentum-based algorithms and with the numerical schemes. These are based upon the idea of adaptive learning rates. The approach of using the operator splitting technique is new in the field of machine learning and one can see that it is an efficient way of developing inertial optimizers. \color{black} Concerning inertial algorithms, we follow [Wilson et. al., 2017], in which one infers that SGD and \ref{NaG} generalize much better than the adaptive algorithms like Adam and Adagrad. Also, even when these adaptive algorithms achieve the same or better training accuracy, the non-adaptive ones give, in the long run, much better accuracy values on the test data. We have observed empirically that different types of inertial algorithms, as in the case of \color{red} SSA1 \color{black} and \color{red} SSA2 \color{black} generalize better than the adaptive optimization methods. At the same time,  in \cite{Keskar}, the authors have developed a method in order to make Adam generalize as good as SGD, that is based on a triggering condition and this represents a key idea in order to make adaptive algorithms to share the same behavior as the inertial ones. This is the first motivation to the fact that we have emphasized the key role of inertial algorithms that are closely related, but ultimately different, than \ref{NaG}. On the other hand, we point out that, in the optimization community, the sequential splitting method has drawn attention due to the recent article of M. Muehlebach and M.I. Jordan, \cite{Jordan}. The authors have showed that Nesterov's method \ref{NaG} is a combination between the sequential splitting method and a symplectic Euler method. Recently, the authors of \cite{Franca} have also used splitting techniques of ODE's, but to proximal-type algorithms. This differs from other papers from the fact that they have used balanced and rebalanced splitting methods that are in deep connection with the equilibrium states of the ODE's subproblems. This represents our second motivation, in the sense that it is a strong argument that operator splitting techniques are not only an elegant approach, but also it gives some insights concerning the evolution equations associated to the discretization methods. Elseways, the comparison of our splitting-based algorithms were done with some classical optimizers, like in the paper \cite{Nadam}, \cite{Adam}, \cite{Wilson}, \cite{Adadelta} and \cite{Sutskever}. \color{black}
Our full implementation is based on Python package \texttt{PyTorch}. In a future work, we aim to add our algorithms to \texttt{Keras} and \texttt{Tensorflow}. We consider adding these algorithms to \texttt{Autograd}, so as to be used as optimizers for particular objective functions, which, in their turn, can be deployed for the approximation of solutions of multi-dimensional initial value problems and boundary value problems. We also intend to present mathematically rigorous proof for the convergence rate of these optimization algorithms for convex and non-convex objective functions. For the convex functions, our aim is to employ the technique concerning suitable Lyapunov functions. Such work is underpinned by \cite{DaSilva}, where Lyapunov functions were used for the associated dynamical systems of some adaptive optimizers. To wrap things up, non-convex optimization functions call for the use of the KL property of the underlying regularization of the objective function, as in \cite{Bot}.

\section*{Acknowledgments}
We would like to thank \texttt{DirectMailers} for allowing us to use their NVIDIA DGX-1 server with eight \texttt{Tesla V100-SXM2} GPUs, which enabled us to do our simulations for the
training of the neural network.  Furthermore, we give our appreciation to Luigi Malag\` o from Romanian Institute of Science and Technology (RIST) from Cluj-Napoca for some valuable suggestions regarding the effect of the learning rate with respect to the splitting-type optimization algorithms.  Finally, we give our appreciation to anonymous reviewers who have helped us organize the present research article, since their comments improved greatly the experimental part of our research.

\end{document}